\documentclass[10pt,journal,compsoc]{IEEEtran}

\usepackage{times}
\usepackage{epsfig}
\usepackage{graphicx}
\usepackage{amsmath}
\usepackage{amssymb}
\usepackage{multirow}
\usepackage{diagbox}
\usepackage{placeins}

\renewcommand{\baselinestretch}{0.95} \normalsize
\newenvironment{DIFnomarkup}{}{}

\DeclareMathOperator{\Tr}{Tr}

\begin{document}

\title{Approximate Fisher Information Matrix to Characterise the Training of Deep Neural Networks}

\begin{DIFnomarkup}
\author{Zhibin Liao\textsuperscript{*}\thanks{This work was partially edited while the first author was a post-doctoral research fellow with the Robotics and Control Laboratory at the University of British Columbia.}  \quad Tom Drummond\textsuperscript{$\dagger$}  \quad Ian Reid\textsuperscript{*} \quad Gustavo Carneiro\textsuperscript{*}\thanks{Supported by Australian Research Council through grants DP180103232, CE140100016 and FL130100102.}\\
Australian Centre for Robotic Vision\\
\textsuperscript{*}University of Adelaide \qquad \textsuperscript{$\dagger$}Monash University\\
{\tt\small \{zhibin.liao, ian.reid, gustavo.carneiro\}@adelaide.edu.au, tom.drummond@monash.edu}
}
\end{DIFnomarkup}

\maketitle

\newcommand{\note}[1]{\textcolor{red}{#1}}

\begin{abstract}

In this paper, we introduce a novel methodology for characterising the performance of deep learning networks (ResNets and DenseNet) with respect to training convergence and generalisation as a function of mini-batch size and learning rate for image classification.  
This methodology is based on novel measurements derived from the eigenvalues of the approximate Fisher information matrix, which can be efficiently computed even for high capacity deep models. 
Our proposed measurements can help practitioners to monitor and control the training process (by actively tuning the mini-batch size and learning rate) to allow for good training convergence and generalisation.
Furthermore, the proposed measurements also allow us to show that it is possible to optimise the training process with a new dynamic sampling training approach that continuously and automatically change the mini-batch size and learning rate during the training process. Finally, we show that the proposed dynamic sampling training approach has a faster training time and a competitive classification accuracy compared to the current state of the art. 

\end{abstract}

\section{Introduction}
\label{sec:introduction}

Deep learning networks (a.k.a. DeepNets), especially the recently proposed deep residual networks (ResNets)~\cite{he2016deep,huang2016deep} and densely connected networks (DenseNets)~\cite{huang2016densely}, are achieving extremely accurate classification performance over a broad range of tasks. 
Large capacity deep learning models are generally trained with  stochastic gradient descent (SGD) methods ~\cite{robbins1951stochastic}, or any of its variants, given that they produce good convergence and generalisation at a relatively low computational cost, in terms of training time and memory usage.
However, a successful SGD training of DeepNets depends on a careful selection of mini-batch size and learning rate, but there are currently no reliable guidelines on how to select these hyper-parameters.

Recently, Keskar et al.~\cite{keskar2016large} proposed numerical experiments to show that large mini-batch size methods converge to sharp minimisers of the objective function, leading to poor generalisation, and small mini-batch size approaches converge to flat minimisers. 
In particular, Keskar et al.~\cite{keskar2016large} proposed a new sensitivity measurement based on an exploration approach that calculates the largest value of the objective function within a small neighbourhood.  
Even though very relevant to our work, that paper~\cite{keskar2016large} focuses only on mini-batch size and does not elaborate on the dynamic sampling training method, i.e., only shows the rough idea of a training algorithm that starts with a small mini-batch and then suddenly switches to a large mini-batch.
Other recent works characterise the loss function in terms of their local minima~\cite{littwin2016loss,soudry2016no,lee2016gradient}, which is interesting but does not provide a helpful guideline for characterising the training procedure.

\begin{figure*}
\begin{center}
\resizebox{\textwidth}{!}{
\begin{tabular}{ccc}
\begin{tabular}{c}
\includegraphics[width=\columnwidth]{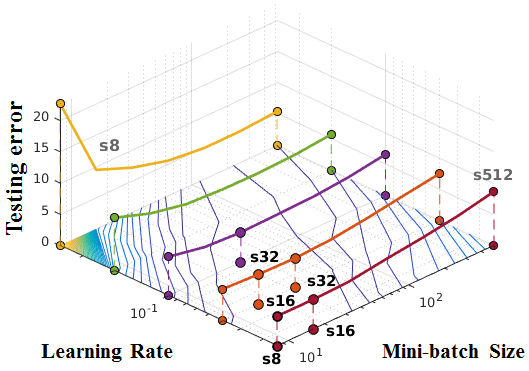}\\
(a)
\\
\includegraphics[width=\columnwidth]{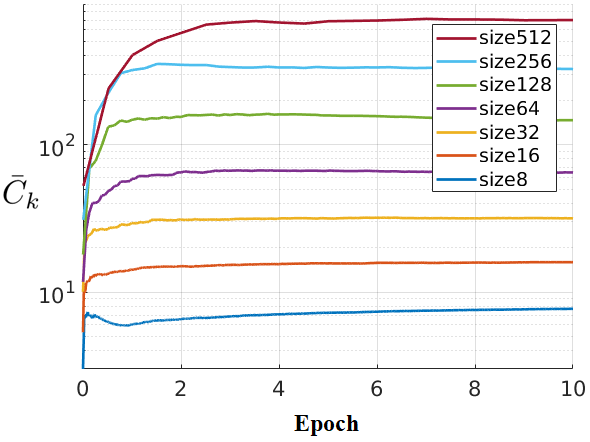} 
\end{tabular}
&
\begin{tabular}{c}
\includegraphics[width=\columnwidth]{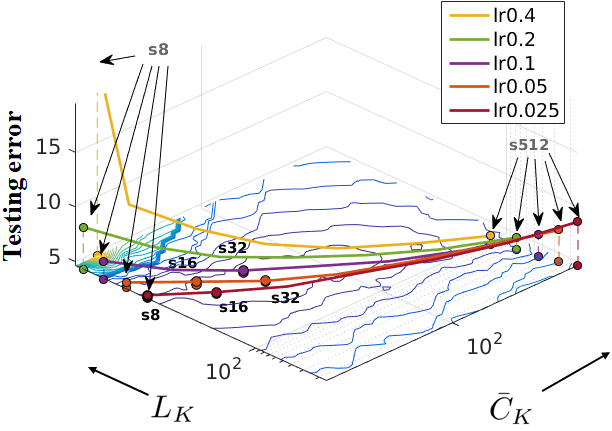} \\
(b)
\\
\includegraphics[width=\columnwidth]{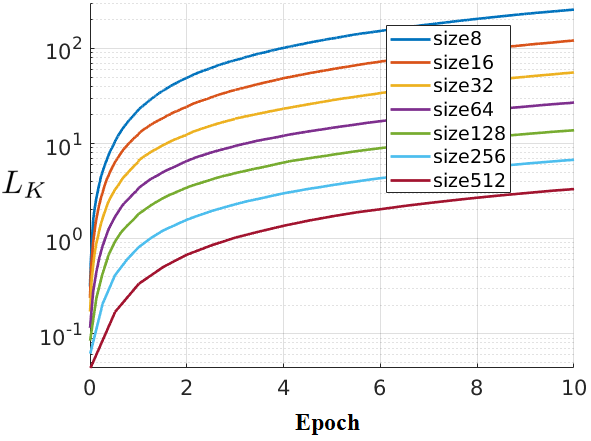}
\end{tabular}
&
\begin{tabular}{c}
\includegraphics[width=\columnwidth]{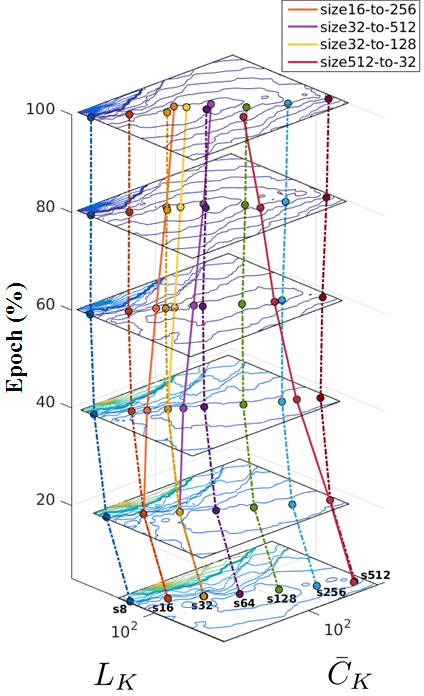}
\end{tabular}
\\
\multicolumn{2}{c}{(c)} & (d) \\
\end{tabular}}
\end{center}
\caption{The evaluation of using different learning rates and mini-batch sizes (a) and corresponding $\bar{C}_K$ and $L_K$ values (b) at the last epoch for the testing set of CIFAR-10~\cite{krizhevsky2009learning}. 
For (a) and (b), the intermediate points (i.e., mini-batch sizes) on each {\bf lr} line are ignored for clarity, except for the top five configurations that produce lowest testing errors.
The stability of the proposed measures over the first 10 epochs is shown in (c), illustrated by a subset of the models with {\bf lr0.1}.
Finally, in (d), the subset of the models in (c) has been used as ``beacons'' to guide the dynamic sampling training (they share the same colors), i.e., we tune the runtime mini-batch size during the training in order to push the $\bar{C}_K$ and $L_K$ values close to the optimum region in order to achieve accurate classification and fast training -- in this example, {\bf s32-to-128} is located closer to the optimum region centre than the other dynamic sampling models: {\bf s16-to-256}, {\bf s32-to-512}, and {\bf s512-to-32} - also {\bf s32-to-128} shows the lowest testing error amongst these four models (hereafter, we use the following notation to represent the model hyper-parameter values: ${\bf s\{\text{{\bf mini-batch size}}\}-lr\{ \text{{\bf learning rate value}} \}}$).}
\label{fig:intro}
\end{figure*}

In this paper, we introduce a novel methodology for characterising the SGD training of DeepNets~\cite{he2016deep, huang2016densely} with respect to mini-batch sizes and learning rate for image classification.  
These experiments are based on the efficient computation of the  eigenvalues of the approximate Fisher information matrix (hereafter, referred to as Fisher matrix)~\cite{chaudhari2016entropy,martens2014new}.  
In general, the eigenvalues of the Fisher matrix can be efficiently computed (in terms of memory and run-time complexities), and they are usually assumed to approximate of the Hessian spectrum~\cite{chaudhari2016entropy,martens2014new, sagun2017empirical, jastrzkebski2017three}, which in turn can be used to estimate the objective function shape.
In particular, Jastrzkebski et al.~\cite{jastrzkebski2017three} show that the Fisher matrix (referred to as the sample covariance matrix in~\cite{jastrzkebski2017three}) approximates well the Hessian matrix when the model is realisable -- that is, when the model's and the training data's conditional probability distributions coincide.  In theory, this happens when the parameter is close to the optimum.
In a deep learning context, this means that the Fisher matrix can be a reasonable approximation of the Hessian matrix at the end of the training (assuming sufficient training has been done), but there is no clear functional approximation to guarantee such approximation through the entire training. 
Nevertheless, in this work we show empirical evidence that the properties of the Fisher matrix can be useful to characterizing the SGD training of DeepNets.

The proposed characterisation of SGD training is based on spectral information derived from the Fisher matrix: 1) the running average of the condition number of the Fisher matrix $\bar{C}_K$  (see \eqref{eq:cumsum_cond} for the definition); and 2) the weighted cumulative sum of the energy of the Fisher matrix $L_K$ (see \eqref{eq:cumsum_laplacian} for the definition).
We observe that $\bar{C}_K$ and $L_K$ enable an empirically consistent characterisation of various models trained with different mini-batch sizes and learning rate.
The motivation of our work is that current hyper-parameter selection procedures rely on validation performance, where the reason why some values are optimal are not well studied, making hyper-parameter selection (particularly on training DeepNets with large-scale datasets) a subjective task that heavily depends on the developer's experience and the ``factory setting'' of the training scripts posted on public repositories.
In Fig.~\ref{fig:intro}-(a), we show an example of hyper-parameter selection with respect to different mini-batch sizes (from 8 to 512) and different learning rates (from $0.025$ to $0.4$, as shown by the lines marked by different colours) for the testing set\footnote{This paper does not pursue the best result in the field, so all models are trained with identical training setup, and we do not try to do model selection using the validation set} of CIFAR-10~\cite{krizhevsky2009learning}, recorded from a trained ResNet model, where the five configurations with the lowest testing errors are highlighted.
In comparison, in Fig.~\ref{fig:intro}-(b) we show the $\bar{C}_K$ and $L_K$ values of the configurations above computed at the final training epoch, showing that the optimal configurations are clustered together in the measurement space and form an optimum region at the centre of the $\bar{C}_K$ vs $L_K$ graph (we define this optimum region to be formed by the level set with minimum error value in the contour plot).
In Fig.~\ref{fig:intro}-(c), we show that the proposed measures are stable in terms of the relative positions of $\bar{C}_K$ and $L_K$ values even during early training epochs, which means that they can be used to predict the performance of new configurations within a few epochs.
From Fig.~\ref{fig:intro}-(c), we can see that regions of performance are formed based on the mini-batch size used in the training process, where relatively small mini-batches tend to produce more effective training process, at the expense of longer training times.  A natural question that can be made in this context is the following: is it possible to reduce the training time with the use of mini-batches of several sizes, and at the same time achieve the classification accuracy of training processes that rely exclusively on small mini-batches?
Fig.~\ref{fig:intro}-(d) shows that the answer to this question is positive, where the proposed $\bar{C}_K$ and $L_K$ values can be used to guide dynamic sampling -- a method that dynamically increases the mini-batch size during the training, by navigating the training procedure in the landscape of $\bar{C}_K$ and $L_K$.  The dynamic sampling approach has been suggested before~\cite{keskar2016large,bottou2016optimization}, but we are unaware of previous implementations.  
Our approach has a faster training time and a competitive accuracy result compared to the current state of the art on CIFAR-10, CIFAR-100~\cite{krizhevsky2009learning}, SVHN~\cite{netzer2011reading}, and MNIST~\cite{lecun1998gradient} using recently proposed ResNet and DenseNet models. 


\section{Literature Review}

In this section, we first discuss stochastic gradient descent (SGD)~\cite{robbins1951stochastic}, inexact Newton and quasi-Newton methods~\cite{fletcher2013practical, bottou2016optimization,byrd1995limited}, as well as (generalized) Gauss-Newton methods~\cite{bertsekas1996incremental,schraudolph2001fast}
the natural gradient method~\cite{amari1998natural}, and scaled gradient iterations such as RMSprop~\cite{tieleman2012lecture} and AdaGrad~\cite{duchi2011adaptive}.  Then we discuss other approaches that rely on numerical experiments to measure key aspects of SGD training~\cite{keskar2016large,littwin2016loss,soudry2016no,lee2016gradient,sagun2016singularity}. 

SGD training~\cite{robbins1951stochastic} is a common iterative optimisation method that is widely used in deep neural networks training.  One of the main goals of SGD is to find a good balance between stochastic and batch approaches to provide a favourable trade-off with respect to per-iteration costs and expected per-iteration improvement in optimising an objective function.  The popularity of SGD in deep learning lies in the tolerable computation cost with acceptable convergence speed.
Second-order methods aim to improve the convergence speed of SGD by re-scaling the gradient vector in order to compensate for the high non-linearity and ill-conditioning of the objective function.  In particular, Newton's method uses the inverse of the Hessian matrix for re-scaling the gradient vector.  This operation has complexity $O(N^3)$ (where $N$ is the number of model parameters, which is usually between $O(10^6)$ and $O(10^7)$ for modern deep learning models), which makes it infeasible.  Furthermore, the Hessian must be positive definite for Newton's method to work, which is not a reasonable assumption for the training of deep learning models.

In order to avoid the computational cost above, several approximate second-order methods have been developed.  For example, the Hessian-free conjugate gradient (CG)~\cite{wright1999numerical} is based on the fact it only needs to compute Hessian-vector products, which can be efficiently calculated with the $\mathcal{R}$-operator~\cite{pearlmutter1994fast} at a comparable cost to a gradient evaluation.  This Hessian-free method has been successfully applied to train neural networks ~\cite{martens2010deep, kiros2013training}.
Quasi-Newton methods (e.g., the BFGS~\cite{fletcher2013practical, bottou2016optimization}) take an alternative route and approximate the inversion of Hessian with only the parameter and gradient displacements in the past gradient iterations.  However, the explicit use of the approximation matrix is also infeasible in large optimisation problems, where the L-BFGS~\cite{byrd1995limited} method is proposed to reduce the memory usage.
The (Generalized) Gauss-Newton method~\cite{bertsekas1996incremental,schraudolph2001fast} approximates Hessian with the Gauss-Newton matrix.
Another approximate second-order method is the natural gradient method~\cite{amari1998natural} that uses the inverse of the Fisher matrix to make the search quicker in the parameters that have less effect on the decision function~\cite{bottou2016optimization}.
Without estimating the second-order curvature, some methods can avoid saddle points and perhaps have some degree of resistance to near-singular curvature~\cite{bottou2016optimization}.
For instance, AdaGrad~\cite{duchi2011adaptive} keeps an accumulation of the square of the gradients of past iterations to re-scale each element of the gradient, so that parameters that have been infrequently updated are allowed to have large updates, and frequently updated parameters can only have small updates.
Similarly, RMSProp~\cite{tieleman2012lecture} normalises the gradient by the magnitude of recent gradients. 
Furthermore, Adadelta~\cite{zeiler2012adadelta} and Adam~\cite{kingma2014adam} improve over AdaGrad~\cite{duchi2011adaptive} by taking more careful gradient re-scaling schemes.

Given the issues involved in the development of (approximate) second-order methods, there has been some interest in the implementation of approaches that could characterise the functionality of SGD optimisation.
Lee et al.~\cite{lee2016gradient} show that SGD converges to a local minimiser rather than a saddle point (with models that are randomly initialised).
Soudry and Carmon~\cite{soudry2016no} provide theoretical guarantees that local minima in multilayer neural networks loss functions have zero training error.
In addition, the exact Hessian of the neural network has been found to be singular, suggesting that methods that assume non-singular Hessian are not to be used without proper modification~\cite{sagun2016singularity}.
Goodfellow et al.~\cite{goodfellow2014qualitatively} found that state-of-the-art neural networks do not encounter significant obstacles (local minima, saddle points, etc.) during the training.
In~\cite{keskar2016large}, a new sensitivity measurement of energy landscape is used to provide empirical evidence to support the argument that training with large mini-batch size converges to sharp minima, which in turn leads to poor generalisation. In contrast, small mini-batch size converges to flat minima, where the two minima are separated by a barrier, but performance degenerates due to noise in the gradient estimation.
Sagun et al.~\cite{sagun2017empirical} trained a network using large mini batches first, followed by the use of smaller mini batches, and their results show that such barrier between these two minima does not exist, so the sharp and flat minima reached by the large and small mini batches may actually be connected by a flat region to form a larger basin. 
Jastrzkebski et al.~\cite{jastrzkebski2017three} found out that the ratio of learning rate to batch size plays an important role in SGD dynamics, and large values of this ratio lead to flat minima and (often) better generalization.
In \cite{smith2017bayesian}, Smith and Le interpret SGD as the discretisation of a stochastic differential equation and predict that an optimum mini-batch size exists for maximizing test accuracy, which scales linearly with both the learning rate and the training set size.
Smith et al.~\cite{smith2017don} demonstrate that decaying learning rate schedules can be directly converted into increasing batch size schedules, and vice versa, enabling training towards large mini-batch size.
Finally in~\cite{goyal2017accurate}, Goyal et al. manage to train in one hour a ResNet~\cite{he2016deep} on ImageNet~\cite{ILSVRCarxiv14} using  mini-batches of size 8K -- this model achieved a competitive result compared to another ResNet trained using mini-batches of size 256.

The dynamic sampling of mini-batch size has been explored in machine learning, where the main focus lies in tuning the mini-batch size in order to improve convergence.
Friedlander and Schmidt~\cite{friedlander2012hybrid} show that an increasing sampling size can maintain the steady convergence rates of batch gradient descent (or steepest decent) methods, and  the authors also prove linear convergence of such method w.r.t. the number of iterations.
In ~\cite{byrd2012sample}, Byrd et al. present a gradient-based dynamic sampling strategy, which heuristically increases the mini-batch size to ensure sufficiently progress towards the objective value descending.
The selection of the mini-batch size depends on the satisfaction of a condition known as the \textit{norm test}, which monitors the norm of the sample variance within the mini-batch.
Simlarly, Bollapragada et al.~\cite{bollapragada2017adaptive} propose an approximate inner product test, which ensures that search directions are descent directions with high probability and improves over the norm test.
Furthermore, Metel~\cite{metel2017mini} presents dynamic sampling rules to ensure that the gradient follows a descent direction with higher probability -- this depends on a dynamic sampling of mini-batch size that reduces the estimated sample covariance.
De et al.~\cite{de2016big} empirically evaluate the dynamic sampling method and observe that it can outperform classic SGD when the learning rate is monotonic, but it is comparable when SGD has fine-tuned learning rate decay.

Our paper can be regarded as a new approach to characterise SGD optimisation, where our {\bf main contributions} are: {\bf1)  new efficiently computed measures derived from the Fisher matrix} that can be used to {\bf explain the training convergence and generalisation of DeepNets with respect to mini-batch sizes and learning rates}, and {\bf 2) a new dynamic sampling algorithm that has a faster training process and competitive classification accuracy} compared to recently proposed deep learning models.

\begin{figure*}
\begin{center}
\resizebox{0.8\textwidth}{!}{%
\begin{tabular}{cc}
\includegraphics[width=0.7\columnwidth]{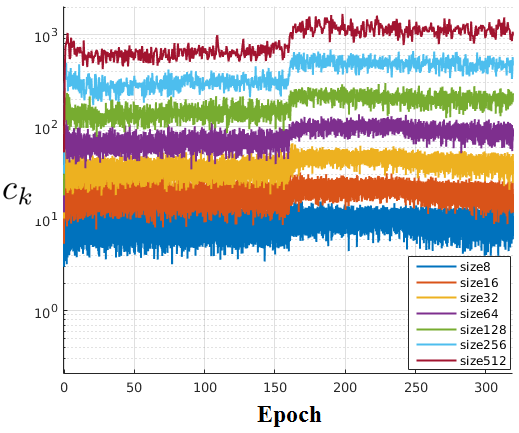} & 
\includegraphics[width=0.7\columnwidth]{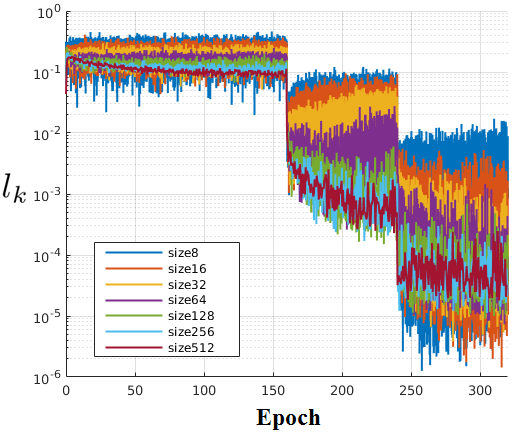} \\
(a) $c_k$ values at epochs $\{1,...,320\}$ &
(b) $l_k$ values at epochs $\{1,...,320 \}$ \\
\includegraphics[width=0.7\columnwidth]{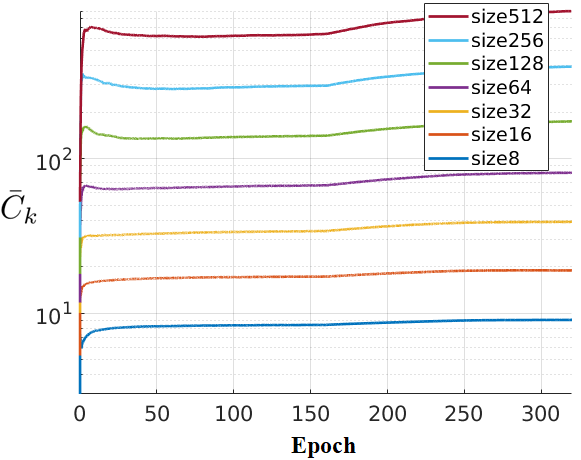} &
\includegraphics[width=0.7\columnwidth]{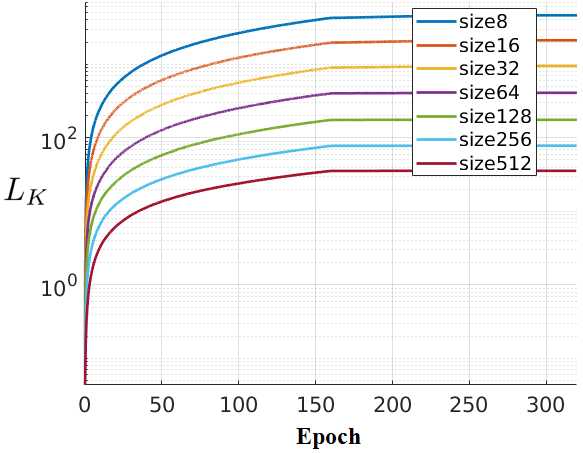}\\
(c) $\bar{C}_k$ values up to epochs $\{ 1,...,320\}$ &
(d) $L_k$ values up to epochs $ \{ 1,...,320 \}$ \\
\end{tabular}%
}
\end{center}
\caption{An illustration of $\{c, l\}_k$ (the first row) and $\{C, L\}_k$ (the second row) of tested ResNet models.  These models are trained on CIFAR-10 with the configurations of mini-batch sizes $ {|\mathcal{B}_k|} = \{8, ..., 512\}$ and initial learning rate $\alpha_1 = 0.1$, $\alpha_k$ is reduced 10 fold at $k \in 161^{\text{st}}$ epoch and $k \in 241^{\text{st}}$ epochs. In general, the sampled measures $\{c,l\}_k$ are noisy, leading to our proposal of the cumulative measures $\{C,L\}_k$ to assess the entire training procedure. The x-axis is shown by epoch instead of iteration to align the readings of different training configurations because the use of small mini-batch sizes increases the number of iterations to complete an epoch.}
\label{fig:cond_eig}
\end{figure*}

\section{Methodology}

In this section, we assume the availability of a dataset $\mathcal{D} = \{ \mathbf{x}_i, y_i \}_{i=1}^{|\mathcal{D}|}$, where the $i^{th}$ image $\mathbf{x}_i:\Omega \rightarrow \mathbb R$ ($\Omega$ denotes image lattice) is annotated with the label $y_i \in \{1,...,C\}$, with $C$ denoting the number of classes.  This dataset is divided into  the following mutually exclusive sets: training $\mathcal{T} \in \mathcal{D}$ and testing $\mathcal{S} \in \mathcal{D}$.

The ResNet model~\cite{he2016deep} is defined by a concatenation of residual blocks, with each block defined by:
\begin{equation}
r_l(\mathbf{v}_l) = f_r(\mathbf{v}_l,\mathbf{W}_l) + \mathbf{v}_l,
\label{eq:residual_block}
\end{equation}
where $l \in \{1,...,L\}$ indexes the residual blocks, $\mathbf{W}_l$ denotes the parameters for the $l^{th}$ block, $\mathbf{v}_l$ is the input, with the image input of the model being represented by $\mathbf{v}_1 = \mathbf{x}$, $f_r(\mathbf{v}_l,\mathbf{W}_l)$ represents a residual unit containing a sequence of linear and non-linear transforms~\cite{nair2010rectified}, and batch normalisation~\cite{ioffe2015batch}.  Similarly, the DenseNet model~\cite{huang2016densely} is defined by a concatenation of dense layers, with each layer defined by:
\begin{equation}
d_l(\mathbf{v}_l) = f_d([\mathbf{v}_1, ..., \mathbf{v}_l],\mathbf{W}_l),
\end{equation}
where $[.]$ represents the concatenation operator, $f_d([...],\mathbf{W}_l)$ contains a sequence of transformations and normalisations similar to $f_r$ of (\ref{eq:residual_block}).

The full model is defined by:
\begin{equation}
f(\mathbf{x}, \theta ) = f_{out} \circ b_L \circ ... \circ b_1( \mathbf{x} ),
\label{eq:baseline_model}
\end{equation}
where $\circ$ represents the composition operator, $b \in \{r, d\}$ represents the choice of computation block, $\theta \in \mathbb R^P$ denotes all model parameters $\{ \mathbf{W}_1, ... \mathbf{W}_L \} \bigcup \mathbf{W}_{out}$, and $f_{out}(.)$ is a linear transform parameterised by weights $\mathbf{W}_{out}$ with a softmax activation function that outputs a value in $[0,1]^C$ indicating the confidence of selecting each of the $C$ classes.  The training of the model in (\ref{eq:baseline_model}) minimises the multi-class cross entropy loss $\ell ( . )$ on the training set $\mathcal{T}$, as follows:
\begin{equation}
\theta^* = \arg \min_{\theta} \frac{1}{|\mathcal{T}|}\sum_{i \in \mathcal{T}} \ell \left ( y_i , f(\mathbf{x}_i, \theta )  \right ).
\label{eq:training_CNN}
\end{equation}

The SGD training minimises the loss in (\ref{eq:training_CNN}) by iteratively taking the following step:
\begin{equation}
\theta_{k+1} = \theta_{k} -   \frac{\alpha_k}{|\mathcal{B}_k|} \sum_{i \in \mathcal{B}_k} \nabla \ell(y_i,f(\mathbf{x}_i,\theta_k)),
\label{eq:sgd}
\end{equation}
where $\mathcal{B}_k$ is the mini-batch for the $k^{th}$ iteration of the minimisation process.
As noted by Keskar et al.~\cite{keskar2016large}, the shape of the loss function can be characterised by the spectrum of the $\nabla^2 \ell(y_i,f(\mathbf{x}_i,\theta_k))$, where sharpness is defined by the magnitude of the eigenvalues. 
However, the loss function sharpness alone is not enough to charaterise SGD training because it is possible, for instance, to adjust the learning rate in order to compensate for possible generalisation issues of the training process~\cite{goyal2017accurate, jastrzkebski2017three, smith2017don}.
In this paper, we combine information derived not only from the spectrum of $\nabla^2 \ell(y_i,f(\mathbf{x}_i,\theta_k))$, but also from the learning rate to characterise SGD training. 
Given that the computation of the spectrum of $\nabla^2 \ell(y_i,f(\mathbf{x}_i,\theta_k))$ is infeasible, we approximate the Hessian by the Fisher matrix (assuming the condition explained in Sec.~\ref{sec:introduction})~\cite{jastrzkebski2017three,chaudhari2016entropy,martens2014new} -- the Fisher matrix is defined by:
\begin{equation}
\mathbf{F}_k = \left ( \nabla \ell(y_{i \in \mathcal{B}_k},f(\mathbf{x}_{i \in \mathcal{B}_k},\theta_k))  \nabla \ell(y_{i \in \mathcal{B}_k},f(\mathbf{x}_{i \in \mathcal{B}_k},\theta_k))^{\top} \right ),
\label{eq:F_k}
\end{equation}
where $\mathbf{F}_k \in \mathbb R^{P \times P}$.

The calculation of $\mathbf{F}_k$ in (\ref{eq:F_k}) depends on the Jacobian $\mathbf{J}_k = \nabla \ell(y_{i \in \mathcal{B}_k},f(\mathbf{x}_{i \in \mathcal{B}_k},\theta_k))$, with $\mathbf{J}_k  \in \mathbb R^{P \times |\mathcal{B}_k|}$.  Given that $\mathbf{F}_k = \mathbf{J}_k \mathbf{J}_k^{\top} \in \mathbb R^{P \times P}$ scales with $P \in [O(10^6),O(10^7)]$ and that we are only interested in the spectrum of $\mathbf{F}_k$, we can compute instead $\widetilde{\mathbf{F}}_k = \mathbf{J}_k^{\top} \mathbf{J}_k \in \mathbb R^{|\mathcal{B}_k| \times |\mathcal{B}_k|}$ that scales with the mini-batch size $|\mathcal{B}_k| \in [O(10^1),O(10^2)]$.  
Note that the rank of $\widetilde{\mathbf{F}}_k$ and $\mathbf{F}_k$ is at most $|\mathcal{B}_k|$, which means that the spectra of $\widetilde{\mathbf{F}}_k$ and $\mathbf{F}_k$ are the same given that both will have at most $|\mathcal{B}_k|$ non-zero eigenvalues.



The {\bf first measure} proposed in this paper is the {\bf running average of the truncated condition number of $\widetilde{\mathbf{F}}_k$} , defined by
\begin{equation}
\bar{C}_K = \frac{1}{K}\sum_{k = 1}^K c_k,
\label{eq:cumsum_cond}
\end{equation}
where $K$ denotes the epoch number, and $c_k = \frac{\sigma_{\text{max}}({\mathcal{E}_k})}{\sigma_{\text{min}}({\mathcal{E}_k})}$ represents the ratio between the largest to the smallest non-zero singular value of $\mathbf{J}_k$ (i.e., we refer to this ratio as the truncated condition number),
with $\mathcal{E}_k$ denoting the set of non-zero eigenvalues computed from $\widetilde{\mathbf{F}}_k$~\cite{wilkinson1965algebraic},  $\sigma_{\text{max}}({\mathcal{E}_k})=\max(\mathcal{E}_k)^{\frac{1}{2}}$, and
$\sigma_{\text{min}}({\mathcal{E}_k})=\min(\mathcal{E}_k)^{\frac{1}{2}}$.  This measure is used to describe the empirical truncated conditioning of the gradient updates observed during the training process.
In Fig.~\ref{fig:cond_eig}-(a) and (c), we show that $c_k$ is a noisy measure and unfit for characterising the training, but $\bar{C}_K$ is more stable, which means that it is able to rank the training procedures more reliably.

The {\bf second measure} is the {\bf weighted cumulative sum of the energy of the Fisher matrix $\widetilde{\mathbf{F}}_k$}, computed by:
\begin{equation}
L_K  = \sum_{k = 1}^K l_k,
\label{eq:cumsum_laplacian}
\end{equation}
where $l_k  = \frac{\alpha_k}{|\mathcal{B}_k|} \left ( \Tr \left ( \widetilde{\mathbf{F}}_k \right ) \right )^{\frac{1}{2}}$,
$\Tr(.)$ represents the trace operator, $\Tr ( \widetilde{\mathbf{F}}_k ) $ approximates the Laplacian, defined by $ \Tr \left ( \nabla^2 \ell(y_i,f(\mathbf{x}_i,\theta_k)) \right ) $, which measures the energy of the approximate Fisher matrix by summing its eigenvalues, and the factor $\frac{\alpha_k}{|\mathcal{B}_k|}$ (with $|\mathcal{B}_k|$ denoting mini-batch size and $\alpha_k$ representing learning rate) is derived from the SGD learning in~\eqref{eq:sgd} -- this factor in~\eqref{eq:cumsum_laplacian} represents a design choice that provides the actual contribution of the energy of the approximate Fisher matrix at the $k^{\text{th}}$ epoch.
Note that in \eqref{eq:cumsum_laplacian}, we apply the square root operator in order to have the magnitude of the values of $L_K$ similar to $\bar{C}_K$ in \eqref{eq:cumsum_cond}.

\subsection{Model Selection}

We observe that deep models trained with different learning rates and mini-batch sizes have values for $\bar{C}_K$ and $L_K$ that are stable, as displayed in Fig.~\ref{fig:intro}-(c) (showing first training epochs) and in Fig.~\ref{fig:cond_eig}-(c,d) (all training epochs).
This means that models and training procedures can be reliably characterised early in the training process, which can significantly speed up the assessment of new models with respect to their mini-batch size and learning rate.
For instance, if a reference model produces a good result, and we know its $\bar{C}_K$ and $L_K$ values for various epochs, then new models must navigate close to this reference model -- see for example in Fig.~\ref{fig:intro}-(b) that {\bf s32-lr0.1} produces good convergence and generalisation with test error $4.78\%\pm0.05\%$ , so new models must try to navigate close enough to it by adjusting the mini-batch size and learning rate.
Indeed, we can see that the two nearest configurations are {\bf s16-0.025} and {\bf s32-lr0.05}, with test error $4.76\%\pm0.11\%$ and $4.67\%\pm0.22\%$ respectively.


\subsection{Dynamic Sampling}
\label{sec:dynamic_sampling}

Dynamic sampling~\cite{friedlander2012hybrid, byrd2012sample} is a method that is believed to improve the convergence rate of SGD by reducing the noise of the gradient estimation with a gradual increase of the mini-batch size over the training process (this method has been suggested before~\cite{friedlander2012hybrid, byrd2012sample}, but we are not aware of previous implementations).  It extends SGD by replacing the fixed size mini-batches $\mathcal{B}_k$ in (\ref{eq:sgd}) with a variable size mini-batch.  
The general idea of this method~\cite{friedlander2012hybrid, byrd2012sample} is that the initial noisy gradient estimations from small mini-batches explore a relatively flat energy landscape without falling into sharp local minima. 
The increase of mini-batch sizes over the training procedure provides a more robust gradient estimation and, at the same time, drives the training into sharper local minima that appear to have better generalisation properties.

With respect to our proposed measures $\bar{C}_K,L_K$ in (\ref{eq:cumsum_cond}),(\ref{eq:cumsum_laplacian}), we notice that dynamic sampling breaks the relative stability between curves of fixed mini-batch sizes, as displayed in Fig.~\ref{fig:intro}-(d).  In general, we note that the application of dynamic sampling allows the curves to move from the region of the original batch size to the region of the final batch size, which means that the training process can be adapted to provide a good trade-off between training speed and accuracy, taking into account that larger mini-batches tend to train faster.  Therefore, we believe that the idea of starting with small and continuously increasing the mini-bathes~\cite{friedlander2012hybrid, byrd2012sample} is just partially true because our results provides evidence not only for such idea, but it also shows that it is possible to start with large mini-batches and continuously decrease them during the training in order to achieve good convergence and generalisation.

\begin{figure*}
\begin{center}
\resizebox{\textwidth}{!}{
\begin{tabular}{cc|cc}
\includegraphics[width=0.25\textwidth]{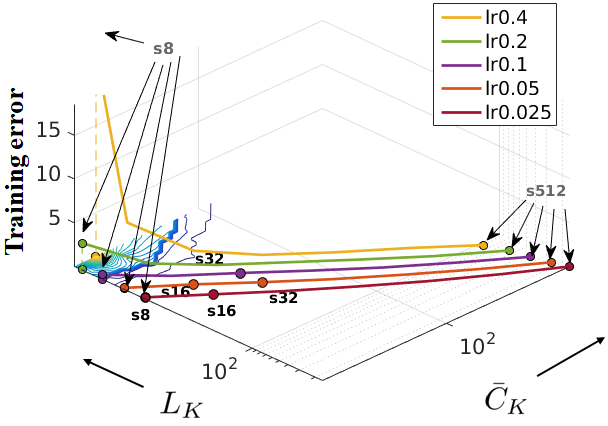} &
\includegraphics[width=0.25\textwidth]{figures_new/cifar10/trans/fig_batchsize_cifar10_test.png} &
\includegraphics[width=0.25\textwidth]{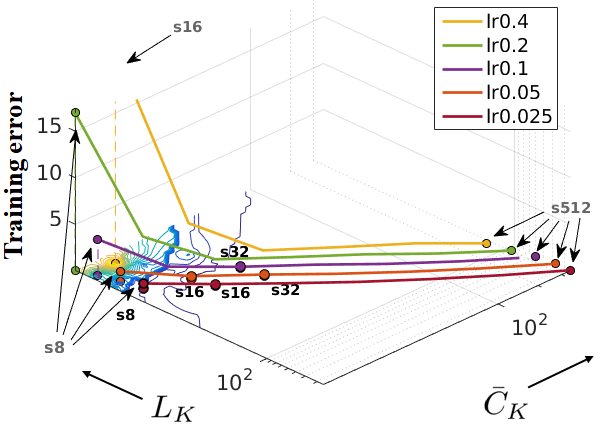} &
\includegraphics[width=0.25\textwidth]{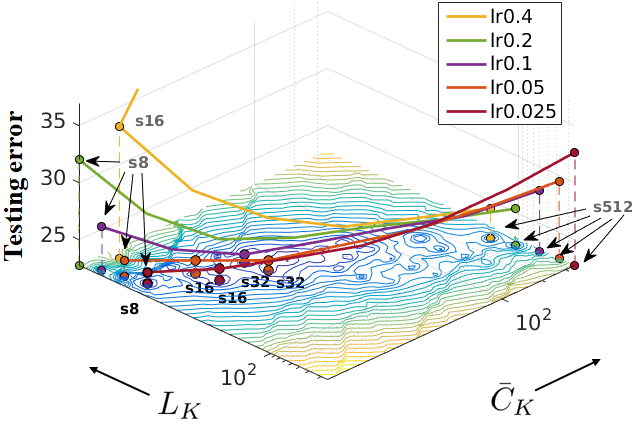} \\
(a) ResNet Training & (b) ResNet Testing & (i) ResNet Training & (j) ResNet Training\\
& & & \\
\includegraphics[width=0.25\textwidth]{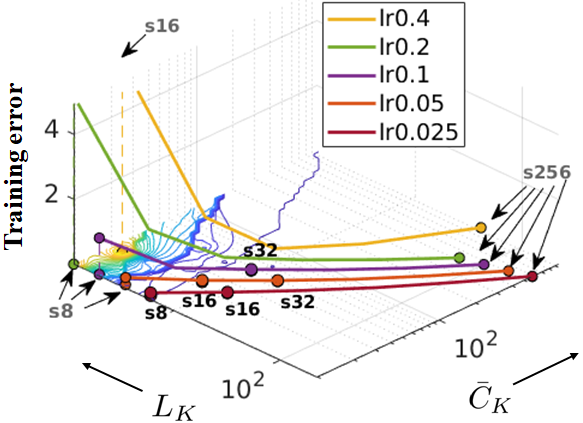} &
\includegraphics[width=0.25\textwidth]{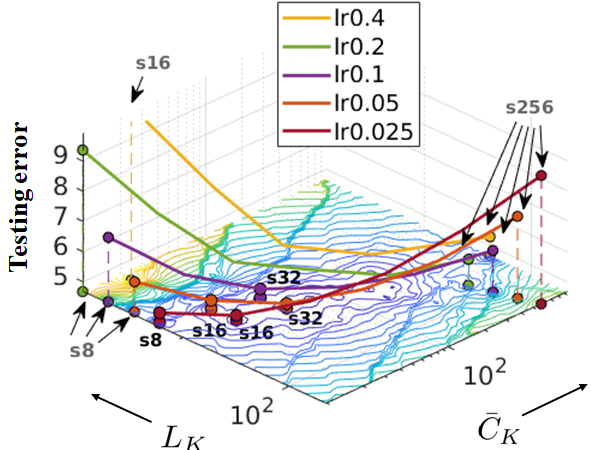} &
\includegraphics[width=0.25\textwidth]{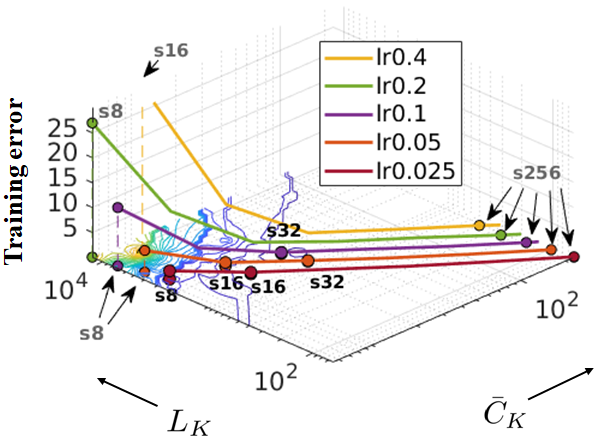} &
\includegraphics[width=0.25\textwidth]{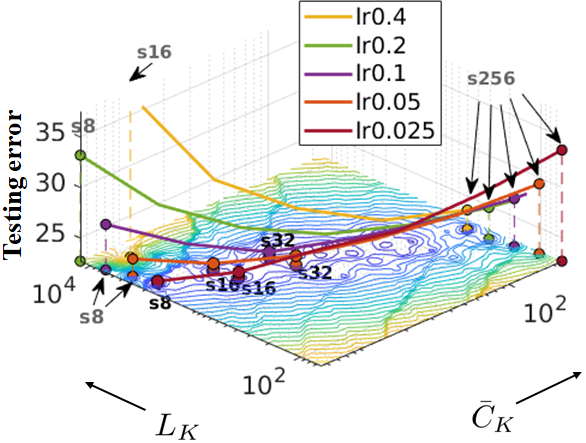} \\
(c) DenseNet Training  & (d) DenseNet Testing & (k) DenseNet Training  & (l) DenseNet Testing \\
& & & \\
& \multicolumn{1}{r|}{CIFAR-10} & \multicolumn{1}{l}{CIFAR-100} & \\
\hline
& \multicolumn{1}{r|}{SVHN} & \multicolumn{1}{l}{MNIST} & \\
& & & \\
\includegraphics[width=0.25\textwidth]{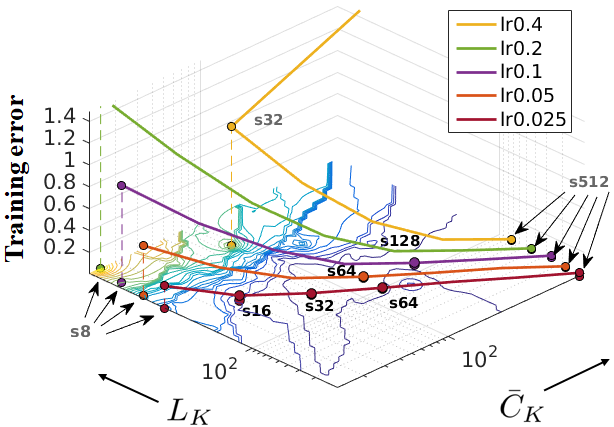} &
\includegraphics[width=0.25\textwidth]{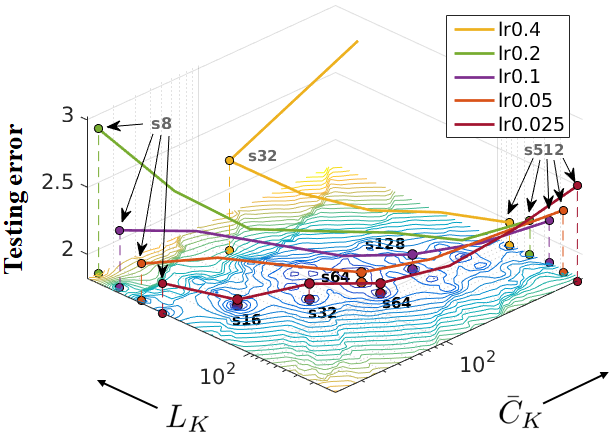} &
\includegraphics[width=0.25\textwidth]{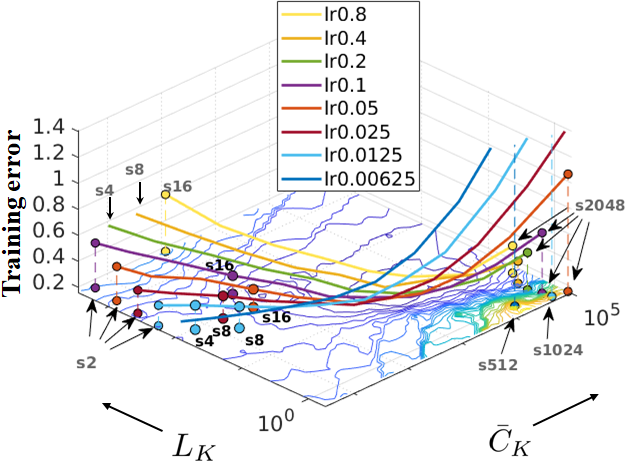} &
\includegraphics[width=0.25\textwidth]{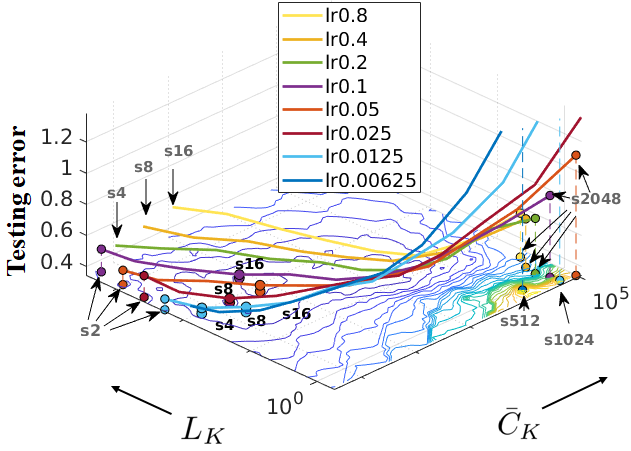} \\
(e) ResNet Training & (f) ResNet Testing & (m) ResNet Training & (n) ResNet Testing \\
& & & \\
\includegraphics[width=0.25\textwidth]{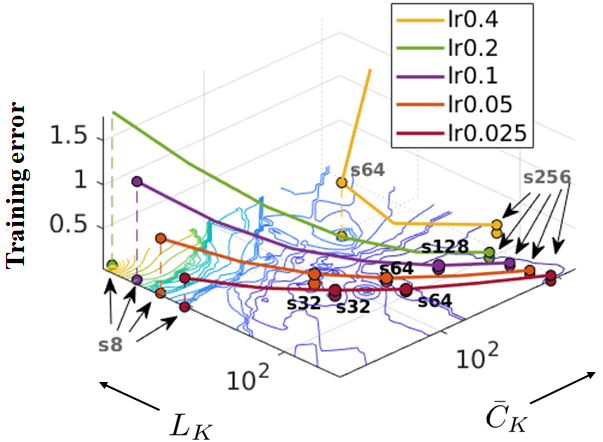} &
\includegraphics[width=0.25\textwidth]{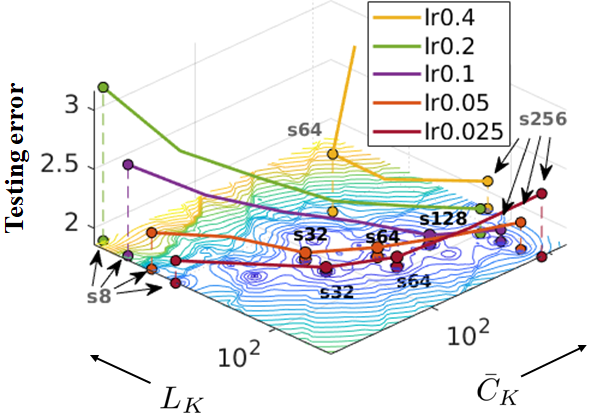} & \includegraphics[width=0.25\textwidth]{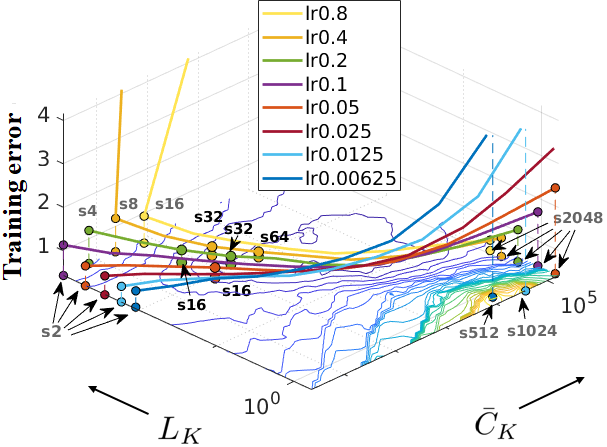}
&
\includegraphics[width=0.25\textwidth]{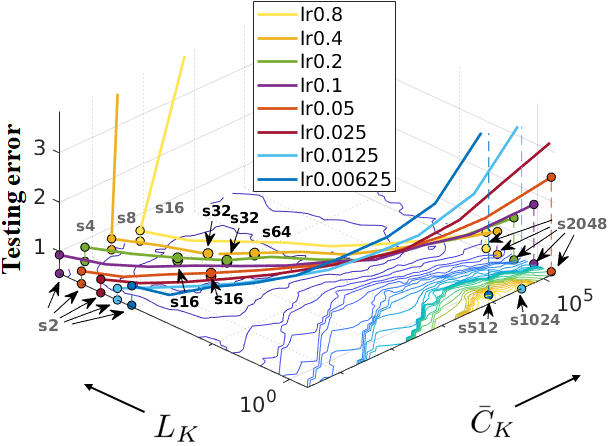} \\
(g) DenseNet Training & (h) DenseNet Testing & (o) DenseNet Training & (p) DenseNet Testing \\
\end{tabular}}
\end{center}
\caption{This graph shows how mini-batch sizes $|\mathcal{B}_k|$ and initial learning rate $\alpha_k$ affect the training performance, which is also related to the proposed measures $\bar{C}_K$ and $L_K$, on four common benchmarks and two model architectures.
We connect the models that use the same $\alpha_k$ value.
Each line omits the intermediate points for clarity except the the top 5 points over all configurations with lowest testing errors.
The gray texts, i.e., {\bf s\{8, 2048\}}, indicate the $|\mathcal{B}_k|$ value at each end of the  $\alpha_k$-connected lines. 
}
\label{fig:batchsize_lr}
\end{figure*}

\section{Experiments}

The experiments are carried out on four commonly evaluated benchmark datasets: CIFAR-10~\cite{krizhevsky2009learning}, CIFAR-100~\cite{krizhevsky2009learning}, SVHN~\cite{netzer2011reading}, and MNIST~\cite{lecun1998gradient}.  
CIFAR-10 and CIFAR-100 datasets contain 60000 $32 \times 32$-pixel coloured images, where 50000 are used for training and 10000 for testing.
SVHN and MNIST are digits recognition datasets where SVHN is a large-scale dataset with over 600000 RGB street view house number plate images and MNIST has 70000 grayscale hand-written digits.

We test our methodology using ResNet~\cite{he2016deep} and DenseNet~\cite{huang2016densely}.
More specifically, we rely on a 110-layer ResNet~\cite{he2016deep} for CIFAR-10/100 and SVHN datasets, including 54 residual units, formed by the following operators in order: $3 \times 3$ convolution, batch normalisation~\cite{ioffe2015batch}, ReLU~\cite{nair2010rectified}, $3 \times 3$ convolution, and batch normalisation.
This residual unit empirically shows better performance than previously proposed residual units (also observed in~\cite{gross2016train} in parallel to our own work).
We use the simplest skip connection with no trainable parameters. 
We also test a DenseNet~\cite{huang2016densely} with 110 layers, involving three stack of 3 dense blocks, where each block contains 18 dense layers. 
By tuning the DenseNet growth rate (i.e., 8, 10 ,14) of each dense block, we manage to composite this DenseNet to have 1.77 million parameters, versus to the 1.73 million in the ResNet.
Due to the simplicity of the MNIST dataset, we use an 8-layer ResNet and an 8-layer DenseNet, including 3 residual units and 6 dense layers (also grouped into 3 dense blocks).
For SGD, we use 0.9 for momentum, and the learning rate decay is performed in multiple steps: 
the initial learning rate is subject to the individual experiment setup, but followed by the same decay policy that decays by $1/10$ at the 50\% training epochs, and by another $1/10$ at 75\% epochs.
That is $161^{st}$ and $241^{st}$ epoch on CIFAR-10/100, and  $21^{st}$ and $31^{st}$ epoch on SVHN and MNIST, where the respective training duration are 320 and 40 epochs.
All training uses data augmentation, as described by He et al.~\cite{he2016deep}.
The scripts for replicating the experiment results are publicly available\footnote{https://github.com/zhibinliao89/fisher.info.mat.torch}.

For each experiment, we measure the training and testing classification error, and the proposed measures $\bar{C}_K$ (\ref{eq:cumsum_cond}) and $L_K$ (\ref{eq:cumsum_laplacian}) -- the reported results are actually the mean result obtained from five independently trained models (each model is randomly initialised).  All experiments are conducted on an NVidia Titan-X and K40 gpus without the multi-gpu computation. 
In order to obtain $\widetilde{\mathbf{F}}_k$ in a efficient manner, the explicit calculation of $\mathbf{J}_k$ is obtained with a modification of the Torch~\cite{torch7} NN and cuDNN libraries (convolution, batch normalisation and fully-connected modules) to acquire the Jacobian $\mathbf{J}_k = \nabla \ell(y_i,f(\mathbf{x}_i,\theta_k)$ during back-propagation.
By default, the torch library calls NVidia cuDNN library in backward training phase to compute the gradient w.r.t the model parameters, where the cuDNN library does not explicitly retain $\mathbf{J}_k$. 
For each type of the aforementioned torch modules, our modification breaks the one batch-wise gradient computation call to $|\mathcal{B}_k|$ individual calls, one for each sample, and then collects the per-sample gradients to form $\mathbf{J}_k$.
Note that the memory complexity to store $\mathbf{J}_k$ scales by $|\mathcal{B}_k|$ times the number of model parameters, which is acceptable for the $1.7$ million parameters in the deep models.
Note that $\mathbf{J}_k$ is formed by iterating over the training samples (in the torch programming layer), which is a slow process given that the underlying NVidia cuDNN library is not open-sourced to be modified to compute the Jacobian within the cuda programming layer directly.
We handle this inefficiency by computing $\widetilde{\mathbf{F}}_k$ at intervals of 50 mini-batches, resulting in a sampling rate of $\approx 2\%$ of training set, so the additional time cost to form $\mathbf{J}_k$ is negligible during the training.
The memory required to store the full $\mathbf{J}_k$ can be reduced by computing $\widetilde{\mathbf{F}}_k = \sum_l \mathbf{J}_{(k,l)}^T \mathbf{J}_{(k,l)}$ for any layer $l$ with trainable model parameters, where $\mathbf{J}_{(k,l)}$ presents the rows of $\mathbf{J}_k$ with respect to the parameters of layer $l$.
This leaves the memory footprint to be only $O(|\mathcal{B}_k|^2)$ for $\widetilde{\mathbf{F}}_k$. 

The training and testing values of the trained models used to plot the figures in this section are listed in the supplementary material. 
At last, due to the higher memory usage of the DenseNet model (compared to ResNet with same configuration), mini-batch size 512 cannot be trained with the 110-layer DenseNet on a single GPU, therefore is excluded from the experiment.

\subsection{Mini-batch Size and Learning Rate}
\label{sec:minibatch_size_and_learning_rate}

In Fig.~\ref{fig:batchsize_lr}, we show our first experiment comparing different mini-batch sizes and learning rates with respect to the training and testing errors, and the proposed measures $\bar{C}_K$ (\ref{eq:cumsum_cond}) and $L_K$ (\ref{eq:cumsum_laplacian}).
The grid values of each 2-D contour map is generated by averaging five nearest error values.
In this section, we refer the error of a model as the final error value obtained when the training procedure is complete.  In general, the main observations for all datasets and models are:
1) each configuration has a unique $\bar{C}_K$ and $L_K$ signature, where no configuration overlays over each other in the space;
2) $|\mathcal{B}_k|$ is directly proportional to $\bar{C}_K$ and inversely proportional to $L_K$;
3) $\alpha_k$ is directly proportional to $\bar{C}_K$ and $L_K$; and
4) small $\bar{C}_K$ and large $L_K$ indicate poor training convergence, and large $\bar{C}_K$ and small $L_K$ show poor generalisation, so the best convergence and generalisation requires a small value for both measures.
Recently, Jastrzkebski et al.~\cite{jastrzkebski2017three} claimed that large $\alpha_k/|\mathcal{B}_k|$ ratio exhibits better generalization property in general.  Our results show that this is true up to a certain value for this ratio.  In particular,  we do observe that the models that produce the top five test accuracy have similar $\alpha_k/|\mathcal{B}_k|$ ratio values (this is clearly shown in the supplementary material), but for very large ratios, when $|\mathcal{B}_k| \in \{8, 16\}$ and $\alpha_k = 0.4$,
then we noticed that convergence issues start to appear.
This is true because beyond a certain increase in the value of $L_K$, $\widetilde{\mathbf{F}}_k$ becomes rank deficient, so that in some epochs (mostly in the initial training period), the smallest eigenvalues get too close to zero, causing some of the $c_k$ values to be large, increasing the value of $\bar{C}_K$ and making the model ill-conditioned.


For both models on {\bf CIFAR-10}, we mark the top five configurations in Fig.~\ref{fig:batchsize_lr}-(a-d) with lowest testing errors, where the best ResNet model is configured by {\bf s32-lr0.05} with $4.67\%\pm0.22\%$, and the best DenseNet is denoted by {\bf s16-lr0.025} with $4.82\%\pm0.07\%$.
This shows that on CIFAR-10, the optimal configurations are with small $|\mathcal{B}_k| \in \{ 8, ..., 32 \}$ and small $\alpha_k \in \{ 0.025, ..., 0.1 \}$.
On {\bf CIFAR-100}, both models show similar results, where the best ResNet model has configuration {\bf s16-lr0.05} with error $23.39\% \pm 0.13\%$, and the best DenseNet is configured as {\bf s8-lr0.025} with error $22.90\% \pm 0.47\%$. 
Note that on CIFAR-100, the optimal configurations are with the same small range of $|\mathcal{B}_k|$ and small $\alpha_k$.
This similarity of the best configurations between CIFAR models is expected because of the similarity in the image data.
We may also conclude that the range of optimal $\bar{C}_K$ and $L_K$ value is not related to the number of classes in the dataset.

Both models show similar results on {\bf SVHN}, where the top ResNet result is reached with {\bf s16-lr0.025} that produced an error of $1.86\%\pm0.03\%$, while the best DenseNet accuracy is achieved by {\bf s32-lr0.025} with error $1.89\%\pm0.01\%$. 
Compared to CIFAR experiments, it is clear that the optimum region on SVHN is ``thinner'', where 
the optimal configurations are with $|\mathcal{B}_k| \in \{ 16, ..., 128 \}$ and $\alpha_k \in \{ 0.025, ..., 0.1 \}$, which appears to shift noticeably towards larger $|\mathcal{B}_k|$ values.
However, compared to the size of the dataset (i.e., SVHN is 10$\times$ larger than CIFARs) such $|\mathcal{B}_k|$ values are still relative small, so that the optimal ratio of $|\mathcal{B}_k|$ with respect to the size of dataset is actually smaller than the ratio observed for the CIFAR experiments.
Note that the errors on SVHN are final testing errors, where we found that the lowest testing error of each individual model  usually occurs between 22 and 25 epochs, and the remaining training gradually overfits the training set, making the final testing error worse by $0.2\%$, on average.
However, we do not truncate the training in order to keep the consistency of training procedures.
Finally, on {\bf MNIST} we test a wider  $|\mathcal{B}_k| \in \{ 2, ..., 2048 \}$ and wider  $\alpha_k \in \{ 0.00625, ..., 0.8 \}$.
The best ResNet model is {\bf s16-lr0.1} with error $0.36\%\pm0.02\%$, and Densenet is {\bf s32-lr0.4} with error $0.54\%\pm0.02\%$.
The optimum region of MNIST on ResNet is with small $|\mathcal{B}_k| \in \{ 4, ..., 16 \}$ and small $\alpha_k \in \{ 0.0125, ..., 0.1 \}$.
On the other hand, the optimum region of MNIST on DenseNet is with slightly large $|\mathcal{B}_k| \in \{ 16, ..., 64 \}$ and large $\alpha_k \in \{ 0.1, ..., 0.4 \}$, showing a divergence on the optimal region w.r.t the architecture.
We emphasis this is due to a large difference in the model parameters, where the 8-layer MNIST ResNet model has 70,000 parameters while the 8-layer DenseNet model has 10,000 parameters.
The main conclusion of this experiment is that the effect of mini-batch size and initial learning rate are tightly related, and can be used to compensate one another to move the $\bar{C}_K$ and $L_K$ values to different places in the measurement space.

\subsection{Functional Relations between Batch Size, Learning Rate, and the Proposed Measures}

\begin{figure*}
\begin{center}
\resizebox{0.9\textwidth}{!}{
\begin{tabular}{cc}
\includegraphics[width=\columnwidth, height=0.55\columnwidth]{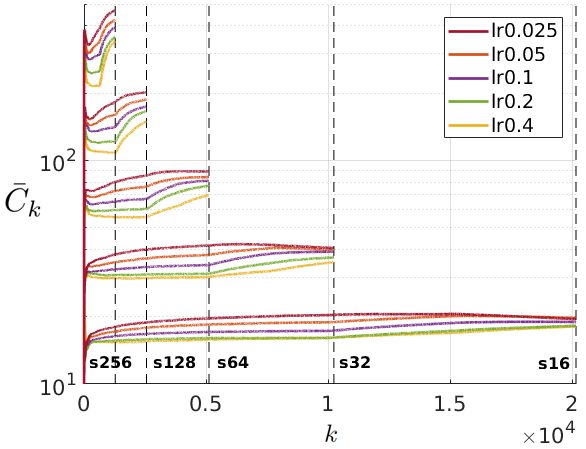} &
\includegraphics[width=\columnwidth, height=0.55\columnwidth]{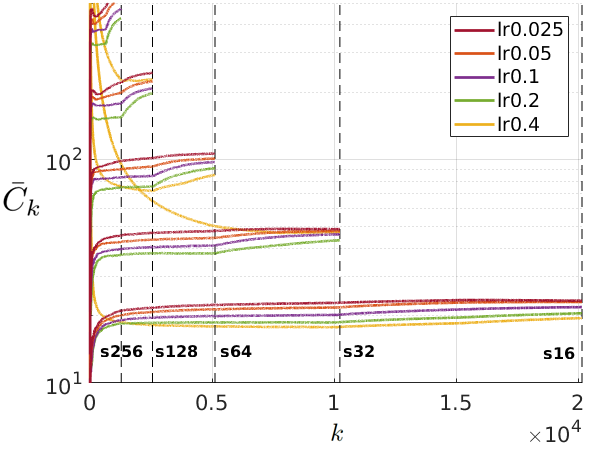} \\
(a) $\bar C_K$ on ResNet training &
(b) $\bar C_K$ on DenseNet training \\
\includegraphics[width=\columnwidth, height=0.55\columnwidth]{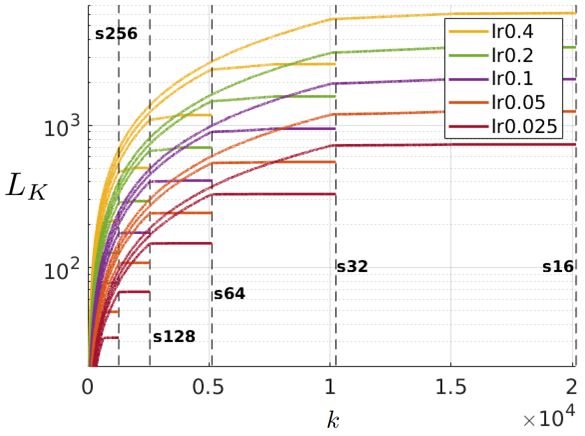} &
\includegraphics[width=\columnwidth, height=0.55\columnwidth]{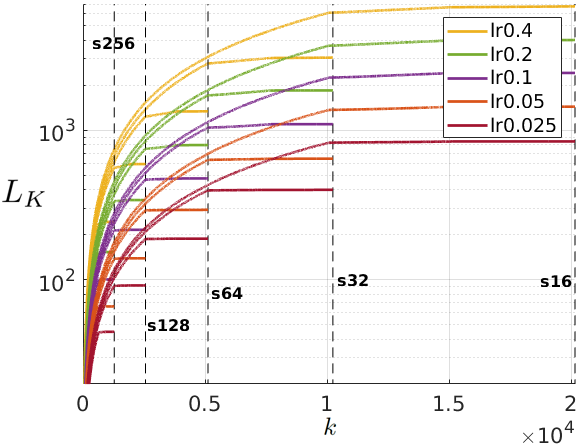} \\
(c) $L_K$ on ResNet training  &
(d) $L_K$ on DenseNet training \\
\end{tabular}}
\end{center}
\caption{The proposed measurements $\bar C_K$ (a,b) and $L_K$ (c,d) for the training of the ResNet (a,c) and DenseNet (b,d) on CIFAR-10, as a function of the number of training iterations (instead of epochs).
The black dotted vertical lines indicate the last iterations for the respective experiments with the same batch size (the results of $s\{8,512\}$ are excluded to avoid a cluttered presentation).  
}
\label{fig:functional_relationship}
\end{figure*}

It is worth noticing the functional relations between hyper-parameters and the proposed measures (due to space restrictions, we only show results for ResNet and DenseNet on CIFAR-10). In Fig.~\ref{fig:functional_relationship}-(a,b), we observe that $\bar C_K$ tends to cluster at similar values for training processes performed with the same mini-batch sizes, independently of the learning rate.
On the other hand, in Fig.~\ref{fig:functional_relationship}-(c,d), we notice that $L_K$ is more likely to cluster at similar values for training processes performed with the same learning rate, independently of the mini-batch size, particularly at the first half of the training.
The major difference between ResNet and DenseNet regarding these functional relations is shown in Fig.~\ref{fig:functional_relationship}-(c), where the learning rate = 0.4 results in poor convergence during the first half of the training process.

\begin{figure*}
\begin{center}
\resizebox{\textwidth}{!}{
\begin{tabular}{cc|cc}
ResNet & DenseNet & ResNet & DenseNet \\
\includegraphics[width=0.7\columnwidth]{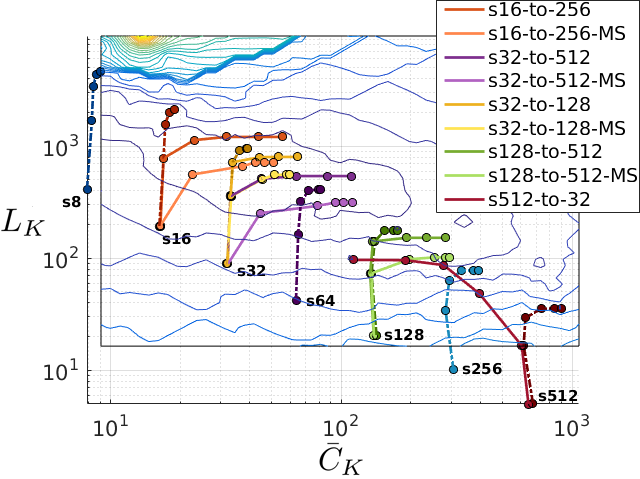} &
\includegraphics[width=0.7\columnwidth]{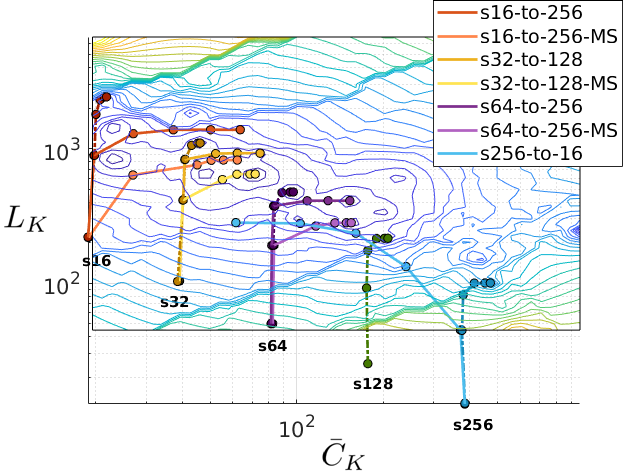} &
\includegraphics[width=0.7\columnwidth]{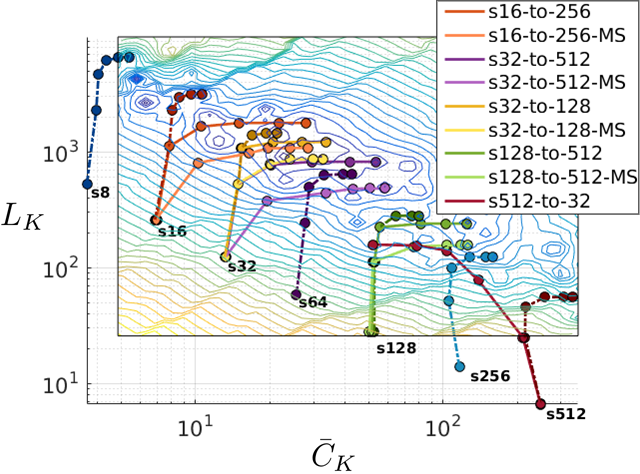} & 
\includegraphics[width=0.7\columnwidth]{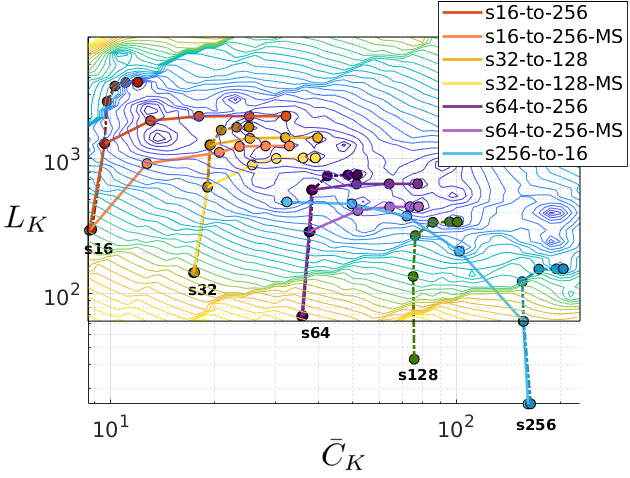} \\
& & & \\
\multicolumn{2}{r|}{(a) CIFAR-10} & \multicolumn{2}{l}{(b) CIFAR-100} \\
\hline
\multicolumn{2}{r|}{(c) SVHN} & \multicolumn{2}{l}{(d) MNIST} \\
& & & \\
ResNet & DenseNet & ResNet & DenseNet \\
\includegraphics[width=0.7\columnwidth]{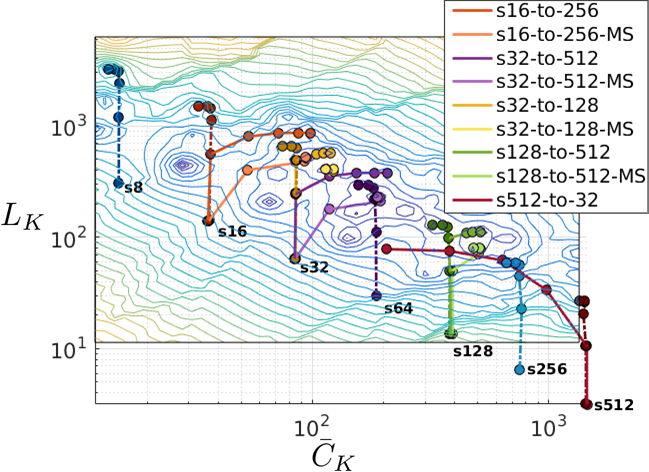} &
\includegraphics[width=0.7\columnwidth]{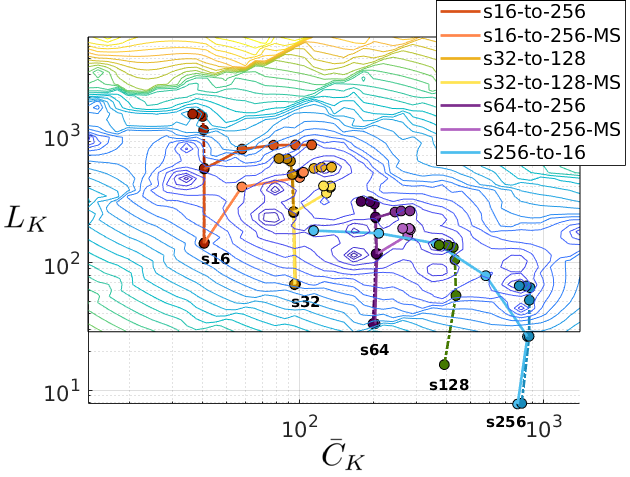} &
\includegraphics[width=0.7\columnwidth]{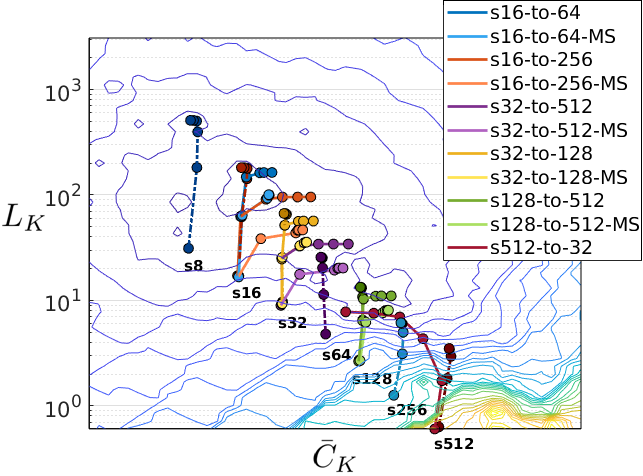} &
\includegraphics[width=0.7\columnwidth]{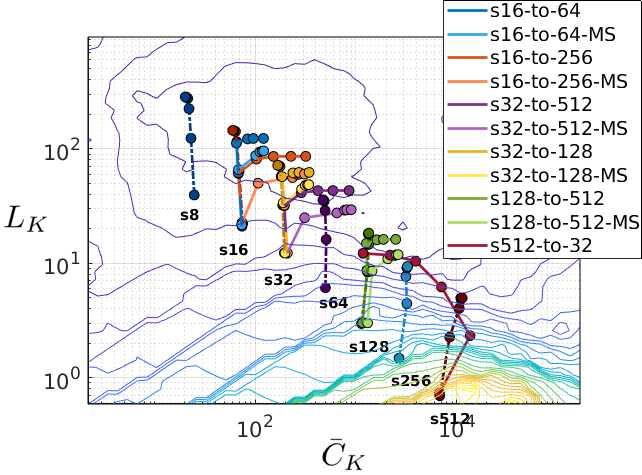} \\
\end{tabular}}
\end{center}
\caption{This graph illustrates the ``travelling history'' of $\bar{C}_K$ and $L_K$ of sevral dynamic sampling models.
Each model is represented by a curve, where the plotted $\bar{C}_K$ and $L_K$ values are extracted from $\{5, 20, 40, 60, 80, 100\}\%$ of the total training epochs, forming six points on each line.
The ``beacon'' models are the corresponding {\bf s\{16, ..., 512\}-lr0.1} models from Fig.~ \ref{fig:batchsize_lr} (for each dataset), which are denoted by dotted curves (not represented in the legend). 
The mini-batch size of each dotted line is marked at the place where the training is at 5\% epochs. 
Furthermore, each dynamic sampling method is designed to share the initial mini-batch size with one of the beacons, so we can observe how they move from the ``roots'' in the graph.
}
\label{fig:dynamic_sampling}
\end{figure*}

\subsection{Dynamic Sampling}

In Fig.~\ref{fig:dynamic_sampling}, we show the runtime analysis of different dynamic sampling alternatives, and how they affect the values of $\bar{C}_K$, $L_K$, as well as the classification errors. 
The dynamic sampling divides the training process into five stages, each with equal number of training epochs and using a particular mini-batch size, i.e., {\bf s32-to-512} uses the mini-batch size sequence \{32, 64, 128, 256, 512\}, 
{\bf s512-to-32} uses \{512, 256, 128, 64, 32\}, 
and {\bf s16-to-64} uses \{16, 16, 32, 32, 64\}.
In each \textbf{dynamic sampling} experiment, the \textbf{first number indicates the initial mini-batch size}, the \textbf{second indicates the final mini-batch size}, and \textbf{-$\emptyset$ or -MS indicates} whether it uses a \textbf{multi-step dynamic sampling approach}.
More specifically, dynamic sampling can be performed over the whole training procedure (indicated by the symbol -$\emptyset$), or within each particular value of learning rate, where the sampling of mini-batch sizes is done over each different learning rate value (denoted by the symbol -MS).
All experiments below use an initial learning rate of 0.1.

{\bf Beacons:}
the ``beacon'' models are the {\bf s\{16, ..., 512\}-lr0.1} models from Fig.~\ref{fig:batchsize_lr}.
In general, the beacon models accumulate $L_K$ faster during the first half of the training procedure than they do during the second half.
On the other hand, the $\bar{C}_K$ measure appears to be more stable during the first half of the training.
However, during the second half of the training, we observe that $\bar{C}_K$ grows on CIFARs (see Fig.~\ref{fig:cond_eig}) but decreases on SVHN and MNIST.

{\bf Dynamic Sampling: }
note in Fig.~\ref{fig:dynamic_sampling} that the dynamic sampling training procedures tend to push $\bar{C}_K$ and $L_K$ away from the initial mini-batch size region towards the final mini-batch size region (with respect to the respective mini-batch size beacons).
Such travel on $L_K$ is faster in the first half of the training procedure than it is in the second half since the growth of $L_K$ is subject to the learning rate (which decays at 50\% and 75\% of the training process).
However, from (\ref{eq:cumsum_cond}), we know that $\bar{C}_K$ is not affected by the learning rate, so it can travel farther towards the final mini-batch size beacon during the second half of training procedure.  For instance, on CIFAR-10 experiment for ResNet, the {\bf s32-to-512} and {\bf s512-to-32} models use the same amount of mini-batch sizes during the training and have the same final $\bar{C}_K$ value but different $L_K$ values.
In Fig.~\ref{fig:dynamic_sampling}-(a) ResNet panel, notice that {\bf s32-to-512} is close to the optimum region, showing a testing error of $5.07\% \pm 0.21\%$, but {\bf s512-to-32} is not as close, showing a testing error of $5.56\% \pm 0.09\%$ for ResNet (the two-sample \textit{t}-test results in a p-value of 0.0037, which means that these two configurations produce statistically different test results, assuming a p-value of 0.05 threshold).  In Fig.~\ref{fig:dynamic_sampling}, we see similar trends for all models and datasets.
Another important observation from Fig.~\ref{fig:dynamic_sampling} is that all dynamic sampling training curves do not lie on any of the curves formed by the beacon models - in fact these dynamic sampling training curves move almost perpendicularly to the beacon models' curves.

\begin{figure*}
\begin{center}
\resizebox{\textwidth}{!}{
\begin{tabular}{ccc}
\includegraphics[width=0.7\columnwidth]{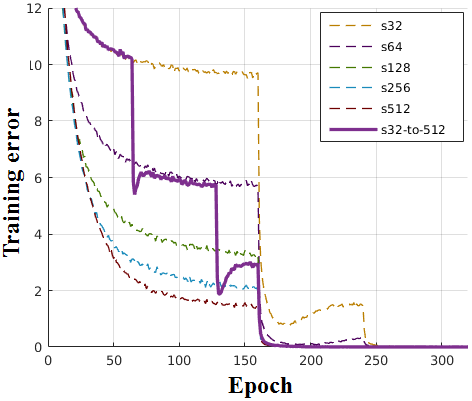} &
\includegraphics[width=0.7\columnwidth]{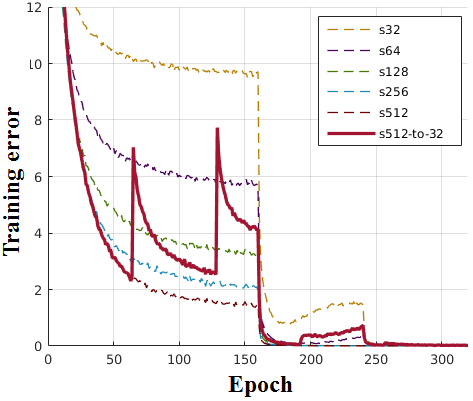} &
\includegraphics[width=0.7\columnwidth]{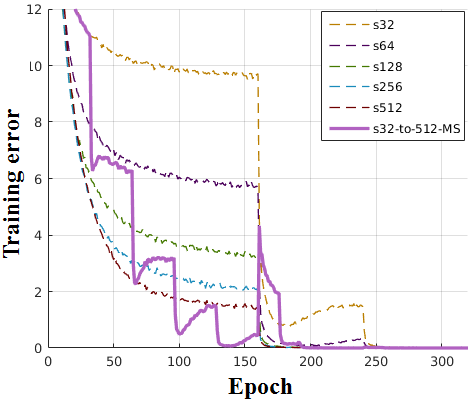} \\
\multicolumn{3}{c}{(a) training error} \\
\includegraphics[width=0.7\columnwidth]{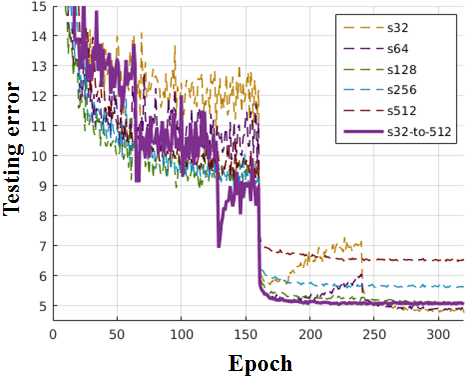} &
\includegraphics[width=0.7\columnwidth]{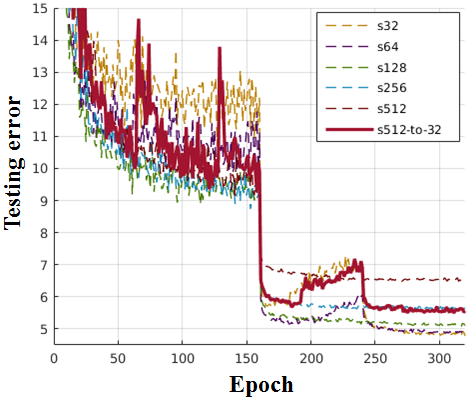} &
\includegraphics[width=0.7\columnwidth]{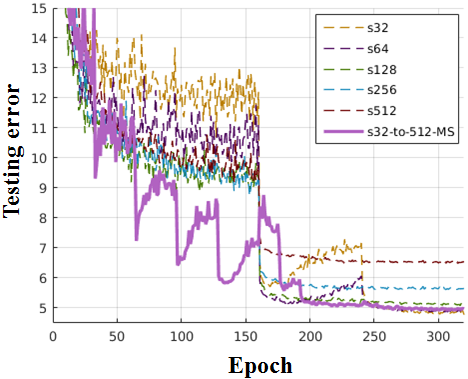} \\
\multicolumn{3}{c}{(b) testing error} \\
\end{tabular}}
\end{center}
\caption{The training (a) and testing error (b) on CIFAR-10 of selected {\bf s32-to-512} (left), {\bf s512-to-32} (centre) dynamic sampling models, and the {\bf s32-to-512-MS} (right) multi-step learning rate decay variant.
}
\label{fig:dynamic_sampling_runtime}
\end{figure*}

We compare the best beacon and dynamic sampling models in Table~\ref{table:dynamic_sampling}.  In general, the results show that dynamic sampling allows a faster training and a similar classification accuracy, compared with the fixed sampling training of the beacons. 
In Fig.~\ref{fig:dynamic_sampling_runtime}, we show the training and testing error curves (as a function of number of epochs) for the {\bf s32-to-512}, {\bf s512-to-32}, and the beacon ResNet models.  Note that the charaterisation of such models using such error measurements is much less stable, when compared with the proposed $\bar{C}_K$ and $L_K$.


\begin{table*}
\begin{center}
\resizebox{\textwidth}{!}{
\begin{tabular}{|c|l|c|c|c||c|c|c|}
\hline
& Model & \multicolumn{3}{c||}{CIFAR-10} & \multicolumn{3}{c|}{CIFAR-100} \\ 
\cline{3-8}
& (best of each)  & {\bf s\#} & {\bf -$\emptyset$} & {\bf -MS} & {\bf s\#} & {\bf -$\emptyset$} & {\bf -MS}  \\
\hline
\multirow{5}{*}{\rotatebox{90}{ResNet}} & Name & 
{\bf s32} & {\bf s32-to-128} & {\bf s32-to-128-MS} &  
{\bf s32} & {\bf s32-to-128} & {\bf s32-to-128-MS} \\ 
\cline{2-8}
& Test Error & 
$4.78\% \pm 0.05\%$ & $4.90\% \pm 0.05\%$ & $4.76\% \pm 0.13\%$ &
$23.46\% \pm 0.21\%$ & $23.90\% \pm 0.31\%$ & $23.69\% \pm 0.34\%$ \\
\cline{2-8}
& p-value vs {\bf s\#} &
\backslashbox{}{} & 0.0048 & 0.72 & 
\backslashbox{}{} & 0.090 & 0.35 \\
\cline{2-8}
& p-value {\bf -$\emptyset$} vs {\bf -MS} & 
\backslashbox{}{} & \multicolumn{2}{c||}{0.029} &
\backslashbox{}{} & \multicolumn{2}{c|}{0.34} \\
\cline{2-8}
& Training Time (h) &
7.7 & 7.0 & 7.1 &
7.9 & 7.1 & 7.1 \\
\hline
\hline
\multirow{5}{*}{\rotatebox{90}{DenseNet}} & Name & 
{\bf s32} & {\bf s64-to-256} & {\bf s32-to-128-MS} &
{\bf s32} & {\bf s32-to-s128} & {\bf s32-to-128-MS} \\
\cline{2-8}
& Test Error & 
$4.96\% \pm 0.12\%$ & $5.03\% \pm 0.03\%$ & $4.63\% \pm 0.10\%$ &
$23.26\% \pm 0.12\%$ & $23.68\% \pm 0.02\%$ & $23.92\% \pm 0.11\%$ \\
\cline{2-8}
& p-value vs {\bf s\#} &
\backslashbox{}{} & 0.38 & 0.022 &
\backslashbox{}{} & 0.048 & 0.0022 \\
\cline{2-8}
& p-value {\bf -$\emptyset$} vs {\bf -MS} &
\backslashbox{}{} & \multicolumn{2}{c||}{0.0029} &
\backslashbox{}{} & \multicolumn{2}{c|}{0.1689} \\
\cline{2-8}
& Training Time (h) & 
16.1 & 14.1 & 15.4 &
16.1 & 14.7 & 14.8 \\
\hline
\hline
& Model & 
\multicolumn{3}{c||}{SVHN} & \multicolumn{3}{c|}{MNIST} \\ 
\cline{3-8}
& (best of each)  & {\bf s\#} & {\bf -$\emptyset$} & {\bf -MS} & {\bf s\#} & {\bf -$\emptyset$} & {\bf -MS} \\
\hline
\multirow{5}{*}{\rotatebox{90}{ResNet}} & Name & 
{\bf s128} & {\bf s128-to-512} & {\bf s32-to-512-MS} &
{\bf s16} & {\bf s16-to-64} & {\bf s16-to-64-MS}\\ 
\cline{2-8}
& Test Error & 
$1.93\% \pm 0.04\%$ & $1.91\% \pm 0.01\%$ & $1.90\% \pm 0.04\%$ &
$0.36\% \pm 0.02\%$ & $0.39\% \pm 0.02\%$ & $0.34\% \pm 0.02\%$ \\
\cline{2-8}
& p-value vs {\bf s\#} & 
\backslashbox{}{} & 0.58 & 0.51 &
\backslashbox{}{} & 0.11 & 0.18 \\
\cline{2-8}
& p-value {\bf -$\emptyset$} vs {\bf -MS} &
\backslashbox{}{} & \multicolumn{2}{c||}{0.69} &
\backslashbox{}{} & \multicolumn{2}{c|}{0.033} \\
\cline{2-8}
& Training Time (h) & 
9.5 & 8.9 & 9.4 &
0.14 & 0.11 & 0.11 \\
\hline
\hline
\multirow{5}{*}{\rotatebox{90}{DenseNet}} & Name &
{\bf s128} & {\bf s64-to-256} & {\bf s64-to-256-MS} &
{\bf s8} & {\bf s32-to-128} & {\bf s16-to-64-MS} \\
\cline{2-8}
& Test Error & 
$1.93\% \pm 0.09\%$ & $2.00\% \pm 0.06\%$ & $2.03\% \pm 0.02\%$ &
$0.61\% \pm 0.06\%$ & $0.57\% \pm 0.03\%$ & $0.60\% \pm 0.01\%$ \\
\cline{2-8}
& p-value vs {\bf s\#} & 
\backslashbox{}{} & 0.31 & 0.16 &
\backslashbox{}{} & 0.43 & 0.78 \\
\cline{2-8}
& p-value {\bf -$\emptyset$} vs {\bf -MS} &
\backslashbox{}{} & \multicolumn{2}{c||}{0.54} &
\backslashbox{}{} & \multicolumn{2}{c|}{0.28} \\
\cline{2-8}
& Training Time (h) & 
21.9 & 20.4 & 20.5 &
0.26 & 0.06 & 0.10 \\
\hline
\end{tabular}%
}
\end{center}
\caption{The comparison between the best beacon model {\bf s\#} (at {\bf lr0.1}), and the best dynamic sampling models {\bf -$\emptyset$}, and {\bf -MS}.}
\label{table:dynamic_sampling}
\end{table*}

\textbf{Dynamic Sampling for Multi-step Learning Rate Decay:}
following the intuition that learning rate decay causes the training process to focus on specific regions of the  energy landscape, we test if the dynamic sampling should be performed within each particular value of learning rate or over all training epochs and decreasing learning rates, as explained above.  This new approach is marked with {\bf -MS} in Fig.~\ref{fig:dynamic_sampling}, where for each learning rate value, we re-iterate through the sequence of mini-batch sizes of the corresponding dynamic sampling policy.
In addition, the training and testing measures for the pair of models {\bf s32-to-512} and {\bf s32-to-512-MS} on ResNet have been shown in leftmost and rightmost columns of Fig.~\ref{fig:dynamic_sampling_runtime} to clarify the difference between the training processes of these two approaches.  Since the new multi-step learning and the original dynamic sampling share the same amount of units of mini-batch sizes during training (which also means that they consume similar amount of training time), then the $\bar{C}_K$ values for both approaches are expected to be similar.  However, the $L_K$ values for the two approaches may have differences, as displayed in Fig.~\ref{fig:dynamic_sampling}.  In general, the {\bf -MS} policy is more effective at limiting the growth of $L_K$ compared to the original dynamic sampling counterpart, pushing the values closer to the optimum region of the graph.
In Table~\ref{table:dynamic_sampling}, we notice that the {\bf -MS} policy produces either significantly better or comparable classification accuracy results, compared to {\bf $\emptyset$} and consumes a similar amount of training time.  Furthermore, both {\bf -MS} and {\bf $\emptyset$} achieve a similar classification accuracy compared to the best beacon models, but with faster training time.

\subsection{Toy Problem}

\begin{figure*}
\begin{center}
\resizebox{\textwidth}{!}{
\begin{tabular}{ccc}
\includegraphics[width=0.7\columnwidth]{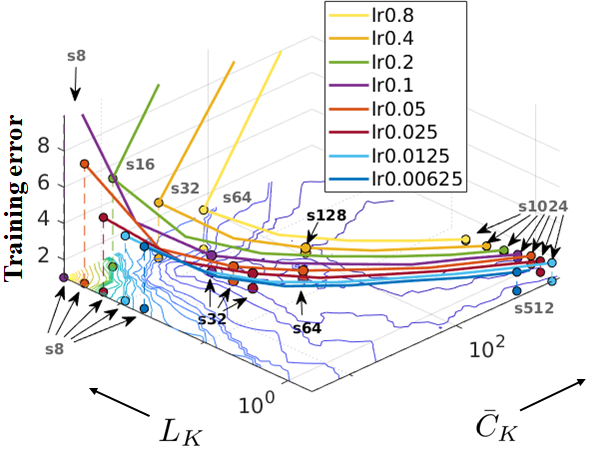} &
\includegraphics[width=0.7\columnwidth]{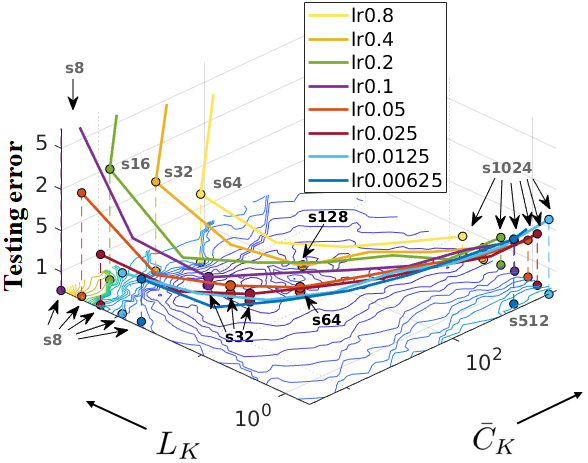} & 
\includegraphics[width=0.7\columnwidth]{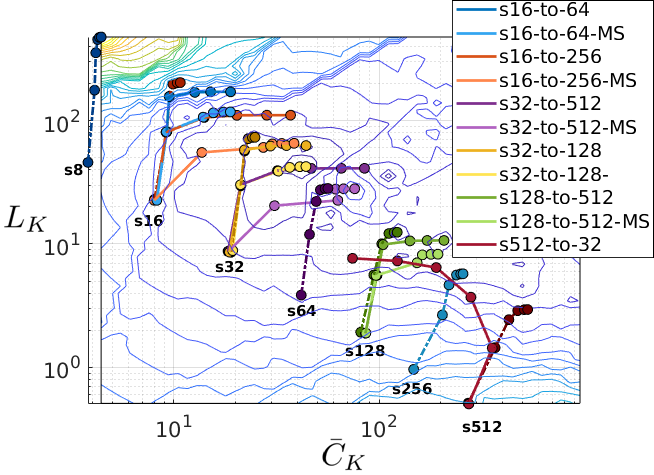} \\
(a) Training & 
(b) Testing & 
(c) Dynamic Sampling History\\
\end{tabular}}
\resizebox{0.7\textwidth}{!}{
\begin{tabular}{cc}
\includegraphics[width=0.7\columnwidth]{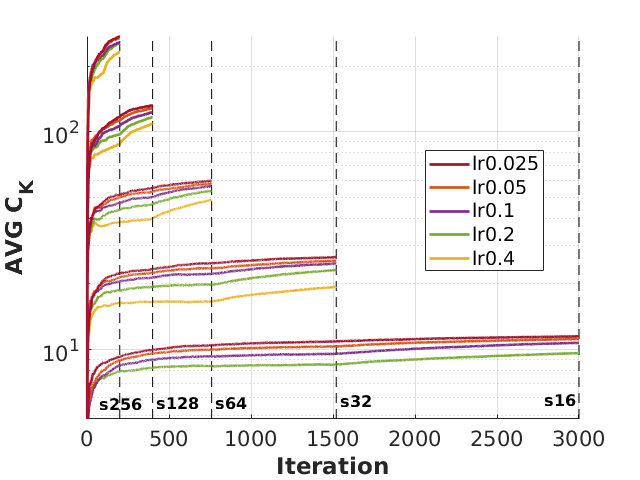} &
\includegraphics[width=0.7\columnwidth]{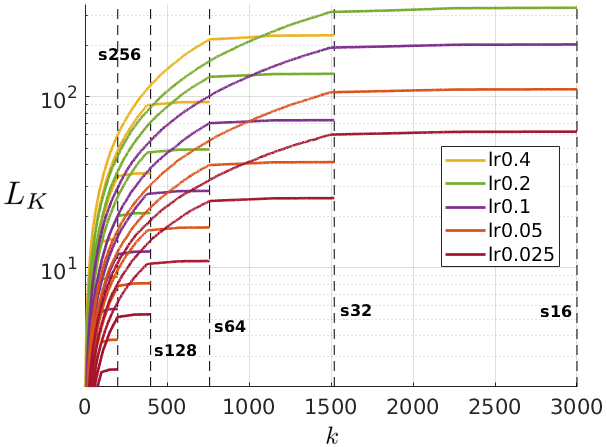} \\
(d) $\bar C_K$ on the training  &
(e) $L_K$ on the training \\
\end{tabular}}
\end{center}
\caption{The proposed measurements quantifying the training of a toy input-500-500-output MLP network.
}
\label{fig:toy_mnist_mlp}
\end{figure*}

In Fig.~\ref{fig:toy_mnist_mlp}, we show that the proposed $\bar{C}_K$ and $L_K$ measurements can also be used to quantify the training of a simple multiple layer perceptron (MLP) network. 
The MLP network has two fully-connected hidden layers, each with 500 ReLU activated nodes. 
The output layer has 10 nodes that are activated with softmax function, allowing it to work as a classifier for the MNIST dataset.
The entire amount of model parameters is 0.77M and the training procedure shares the same training hyper-parameter settings (i.e., momentum and learning rate schedule) of the ResNet and DenseNet models. 
It can be observed from Fig.~\ref{fig:toy_mnist_mlp} that the relative $\bar{C}_K$ and $L_K$ readings for each pair of learning rate and mini-batch size combination is similar to the their counterparts in the ResNet and DenseNet experiments.

\section{Discussion and Conclusion}

The take-home message of this paper is the following: training deep networks, and in particular ResNets and DenseNets, is still an art, but the use a few efficiently computed measures from SGD can provide substantial help in the selection of model parameters, such as learning rate and mini-bath sizes, leading to good training convergence and generalisation.
One possible way to further utilize the proposed $\bar{C}_K$ and $L_K$ in order to achieve a good balance between training convergence and generalisation is to dynamically tune batch size and learning rate so that the $\bar{C}_K$ and $L_K$ measurements do not increase too quickly because this generally means that the training process left the optimal convergence/generalisation region.

In conclusion, we proposed a novel methodology to characterise the performance of two commonly used DeepNet architectures regarding training convergence and generalisation as a function of mini-batch size and learning rate.  
This proposed methodology defines a space that can be used for guiding the training of DeepNets, which led us to propose a new dynamic sampling training approach.  
We believe that the newly proposed measures will help researchers make important decisions about the DeepNets structure and training procedure.  
We also expect that this paper has the potential to open new research directions on how to assess and predict top performing DeepNets models with the use of the proposed measures ($\bar{C}_K$ and $L_K$) and perhaps on new measures that can be proposed in the future.


{\small
\bibliographystyle{IEEEtran}
\bibliography{Bibliography}
}

\newpage
\onecolumn

\begin{center}
\textup{\Huge Approximate Fisher Information Matrix to Characterise the Training of Deep Neural Networks - Supplementary Material}
\end{center}

\vspace{20mm}
\setcounter{section}{0}

\section{Plotting Graphs with Different Timescales}

\begin{figure*}[b]
    \centering
    \resizebox{\textwidth}{!}{%
    \begin{tabular}{cc}
    \includegraphics[width=0.48\columnwidth]{figures_new/cifar10/trans/fig_batchsize_cifar10_test.png} & 
    \includegraphics[width=0.52\columnwidth]{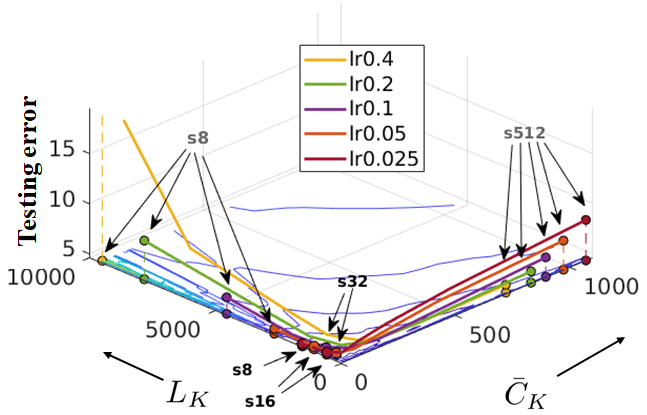} \\
    (a) Log-scale View & (b) Linear-scale View \\
    \end{tabular}%
    }
    \caption{Testing error as a function of $\bar C_K$ and $L_K$ shown in (a) log scale and (b) linear scale, for ResNet on CIFAR-10 dataset. The optimum region is marked by the top-5 test accuracy.}
    \label{fig:axis_scale_comparison}
\end{figure*}

Fig.~\ref{fig:axis_scale_comparison} shows a comparison between the log-scale and linear-scale plots of the testing error as a function of $\bar C_K$ and $L_K$, where Fig.~\ref{fig:axis_scale_comparison}-(a) is identical to Fig.~3-(b) (from main manuscript) that uses log scale for $\bar C_K$ and $L_K$.
This figure suggests that the training processes outside the optimal region can rapidly increase the values of $\bar C_K$ or $L_K$. 
Therefore, it is possible for a practitioner to control the training process to remain in an optimal region by tuning $|\mathcal{B}_k|$ and $\alpha_k$ in order to keep $\bar C_K$ and $L_K$ at relatively low values -- this guarantees a good balance between convergence and generalisation.

\begin{figure*}
\begin{center}
\resizebox{0.75\textwidth}{!}{%
\begin{tabular}{cc}
\includegraphics[width=0.45\columnwidth]{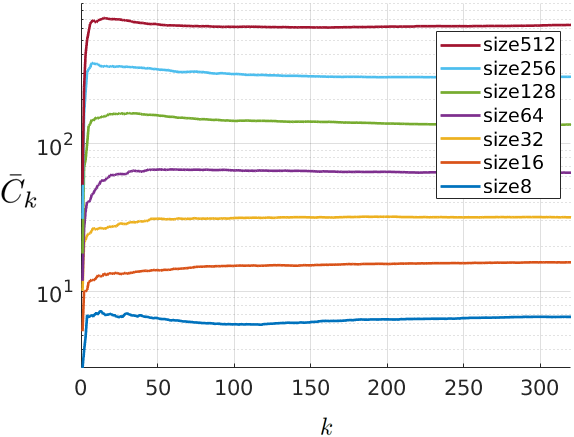} &
\includegraphics[width=0.45\columnwidth]{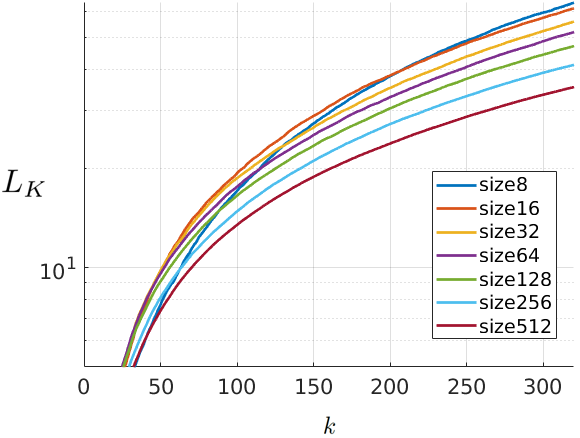}\\
(a) $\bar{C}_k$ values up to 320 iterations &
(b) $L_k$ values up to 320 iterations \\
\includegraphics[width=0.45\columnwidth]{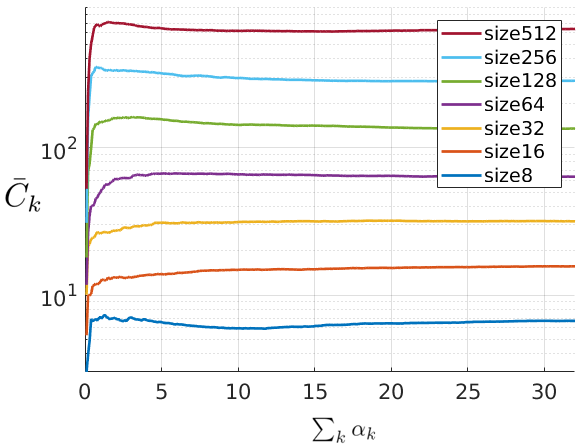} &
\includegraphics[width=0.45\columnwidth]{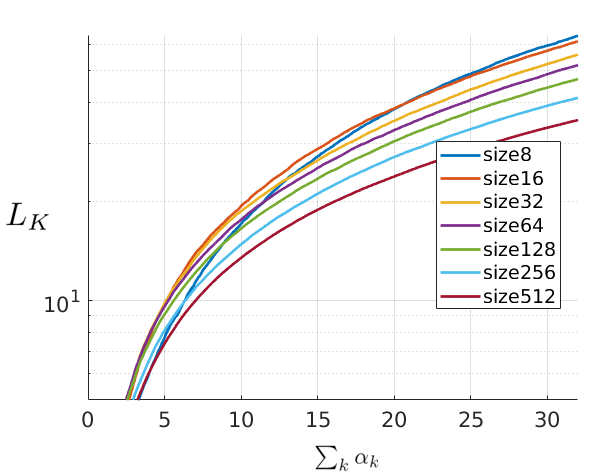}\\
(c) $\bar{C}_k$ values up to iteration $k$ and aligned by $\sum_{k=1}^{320} \alpha_k$ &
(d) $L_k$ values up to iteration $k$ and aligned by $\sum_{k=1}^{320} \alpha_k$ \\
\end{tabular}%
}
\end{center}
\caption{The re-plot of Fig.~1-(c) from the main manuscript, showing $\bar C_k$ (a,c) and $L_k$ (b,d) as a function of $k$ in (a,b), and $\sum_k \alpha_k$ in (c,d).}
\label{fig:intro_supp}
\end{figure*}

\begin{figure*}
\begin{center}
\resizebox{0.75\textwidth}{!}{%
\begin{tabular}{cc}
\includegraphics[width=0.45\columnwidth]{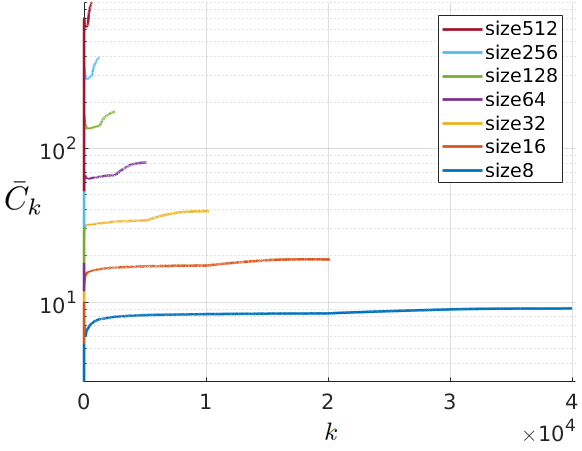} &
\includegraphics[width=0.45\columnwidth]{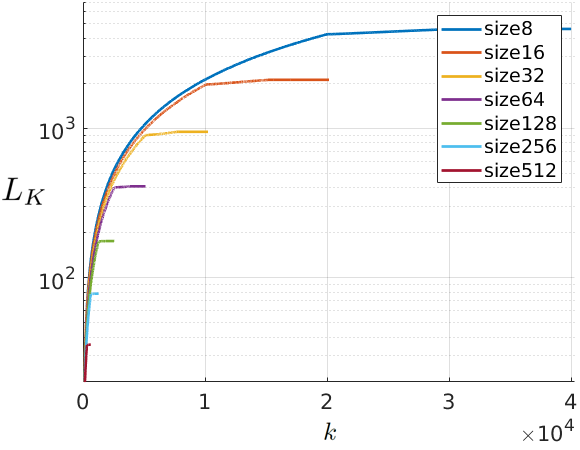}\\
(a) $\bar{C}_k$ values up to iteration $k \in \{ 1,..., \sum_{e=1}^{320} \frac{|\mathcal{T}|}{|\mathcal{B}_e|}\}  $ &
(b) $L_k$ values up to iteration $k \in \{ 1,...,\sum_{e=1}^{320} \frac{|\mathcal{T}|}{|\mathcal{B}_e|} \} $ \\
\includegraphics[width=0.45\columnwidth]{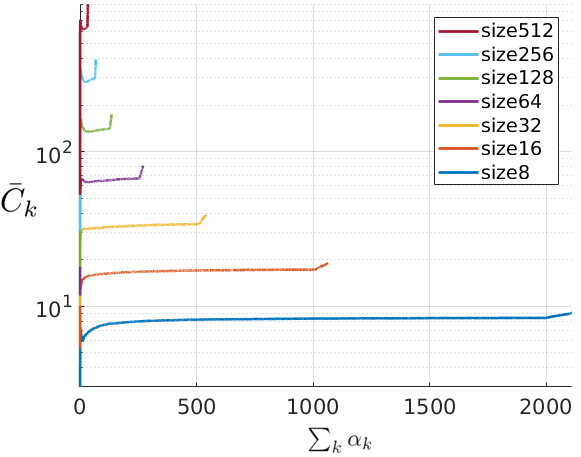} &
\includegraphics[width=0.45\columnwidth]{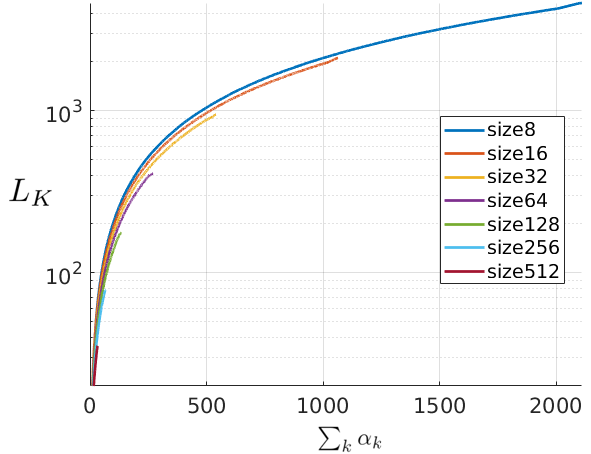}\\
(c) $\bar{C}_k$ values up to iteration $k$ and aligned by $\sum_k \alpha_k$ &
(d) $L_k$ values up to iteration $k$ and aligned by $\sum_k \alpha_k$ \\
\end{tabular}%
}
\end{center}
\caption{Re-plot of Fig.~2-(c,d) from the main manuscript, showing $\bar C_k$ and $L_k$ as a function of the iteration $k$ in (a,b) and $\sum_k \alpha_k$ in (c,d).
}
\label{fig:cond_eig_supp}
\end{figure*}

\begin{figure*}
\begin{center}
\resizebox{0.8\textwidth}{!}{
\begin{tabular}{cc}
\includegraphics[width=0.5\columnwidth]{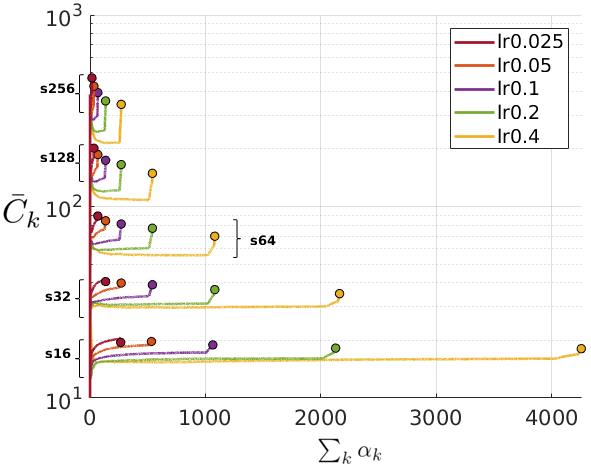} &
\includegraphics[width=0.5\columnwidth]{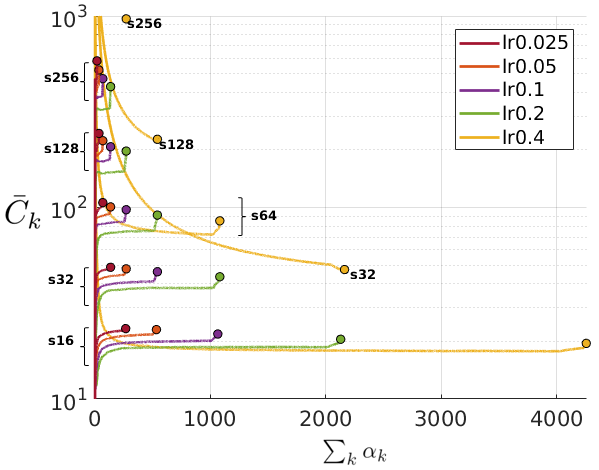} \\
(a) $\bar C_K$ on ResNet training &
(b) $\bar C_K$ on DenseNet training \\
\includegraphics[width=0.5\columnwidth]{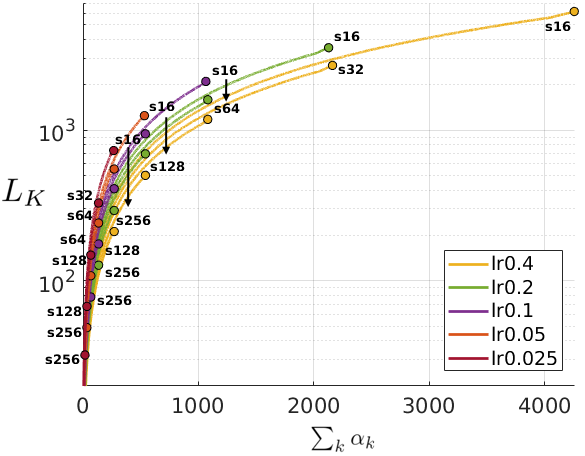} &
\includegraphics[width=0.5\columnwidth]{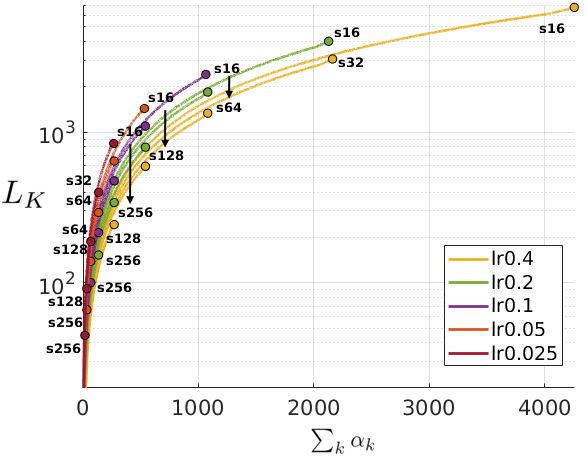} \\
(c) $L_K$ on ResNet training  &
(d) $L_K$ on DenseNet training \\
\end{tabular}}
\end{center}
\caption{The re-plot of Fig~5 from the main article, showing $\bar C_k$ and $L_k$ as a function of the cumulative learning rate, i.e., $\sum_k \alpha_k$ during the training. }
\label{fig:functional_relationship_supp}
\end{figure*}

\begin{figure*}
\begin{center}
\resizebox{\textwidth}{!}{
\begin{tabular}{ccc}
\includegraphics[width=0.3\columnwidth]{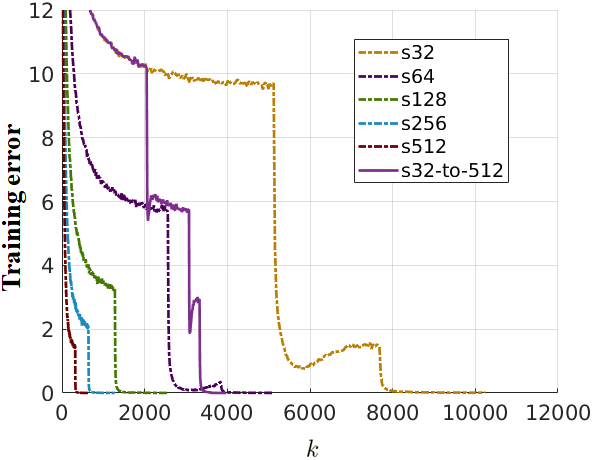} &
\includegraphics[width=0.3\columnwidth]{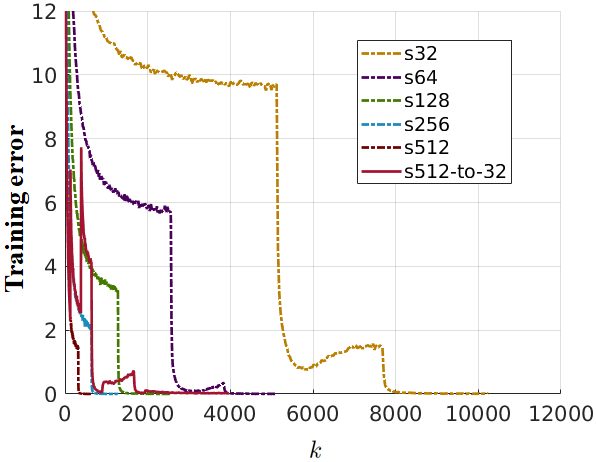} &
\includegraphics[width=0.3\columnwidth]{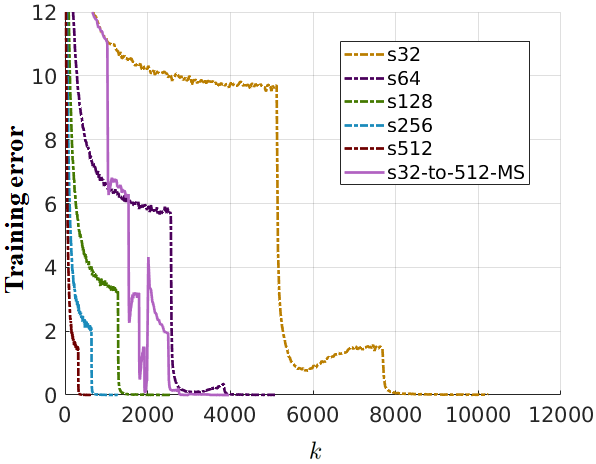} \\
\multicolumn{3}{c}{(a) training error} \\
\includegraphics[width=0.3\columnwidth]{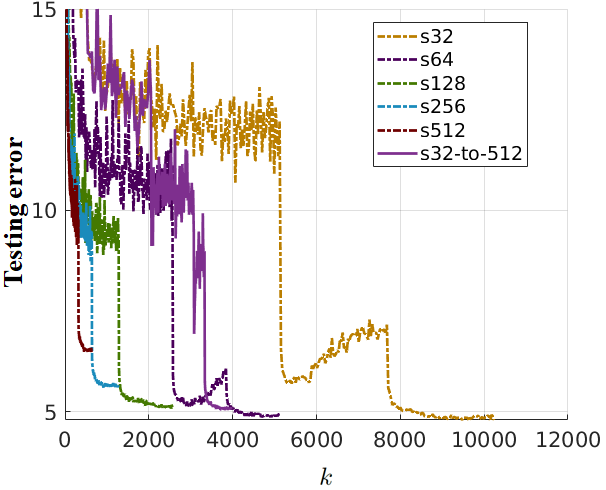} &
\includegraphics[width=0.3\columnwidth]{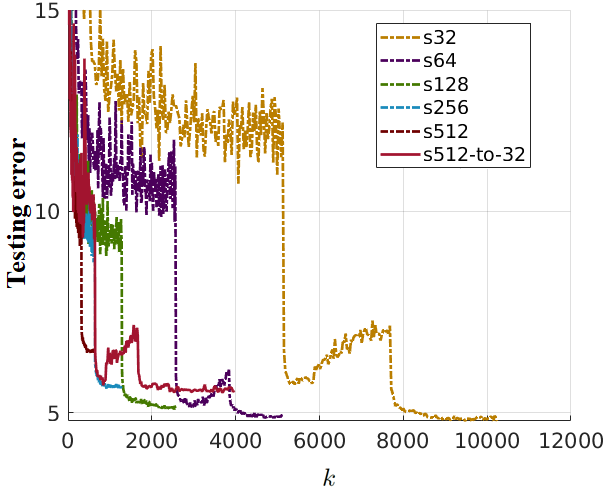} &
\includegraphics[width=0.3\columnwidth]{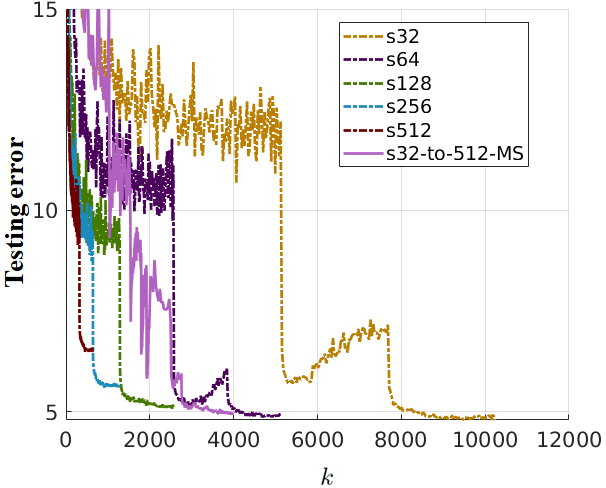} \\
\multicolumn{3}{c}{(b) testing error} \\
\end{tabular}}
\end{center}
\caption{The re-plot of Fig.6 from the main article, showing training and testing errors as a function of the training iteration $k$.}
\label{fig:dynamic_sampling_runtime_iter}
\end{figure*}

\begin{figure*}
\begin{center}
\resizebox{\textwidth}{!}{
\begin{tabular}{ccc}
\includegraphics[width=0.3\columnwidth]{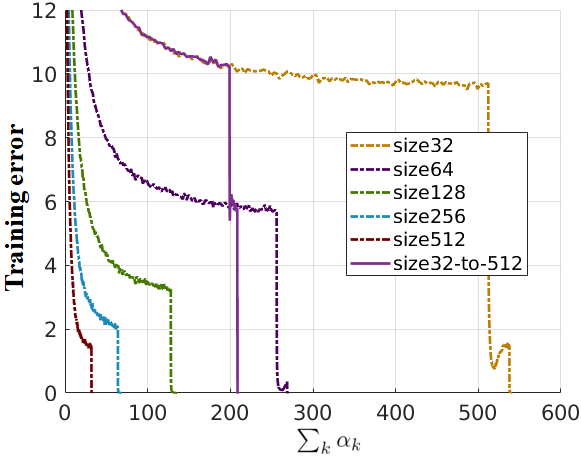} &
\includegraphics[width=0.3\columnwidth]{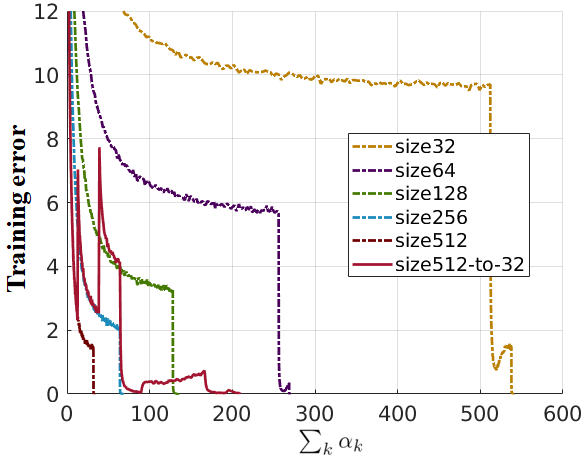} &
\includegraphics[width=0.3\columnwidth]{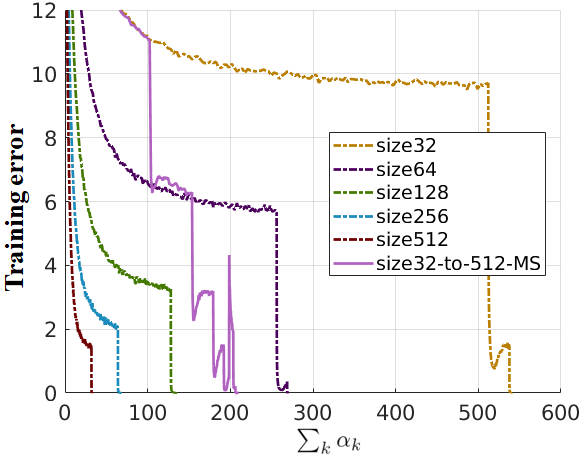} \\
\multicolumn{3}{c}{(a) training error} \\
\includegraphics[width=0.3\columnwidth]{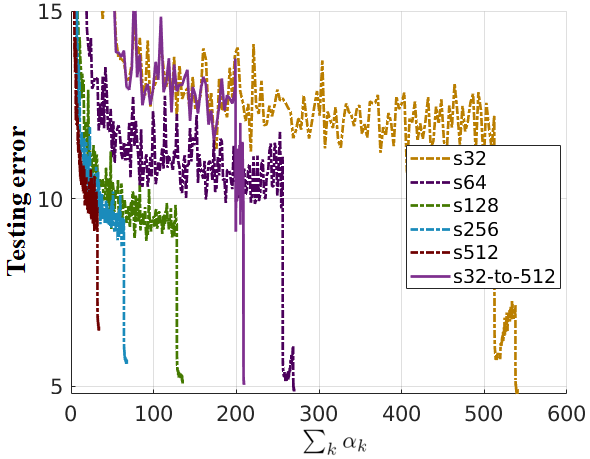} &
\includegraphics[width=0.3\columnwidth]{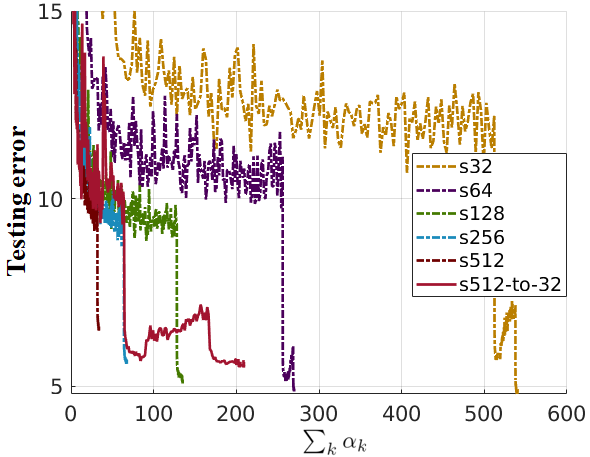} &
\includegraphics[width=0.3\columnwidth]{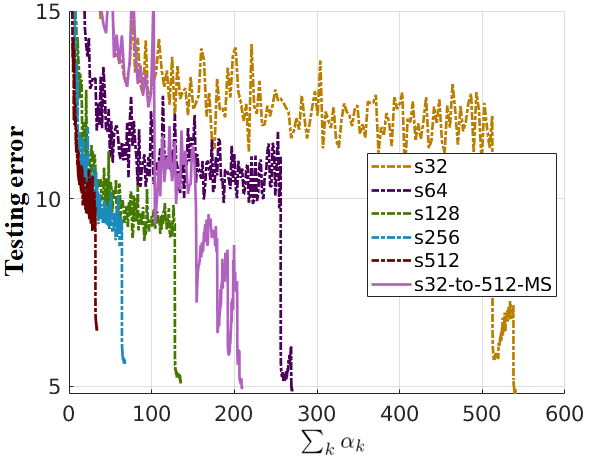} \\
\multicolumn{3}{c}{(b) testing error} \\
\end{tabular}}
\end{center}
\caption{The re-plot of Fig.6 from the main article, showing training and testing errors as a function of the cumulative learning rate, i.e., $\sum_k \alpha_k$.}
\label{fig:dynamic_sampling_runtime_iterlr}
\end{figure*}

\begin{figure*}
\begin{center}
\resizebox{\textwidth}{!}{
\begin{tabular}{cc}
\includegraphics[width=0.5\columnwidth]{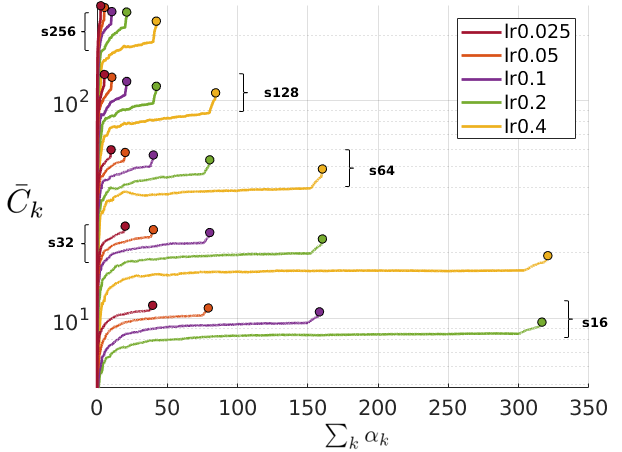} &
\includegraphics[width=0.5\columnwidth]{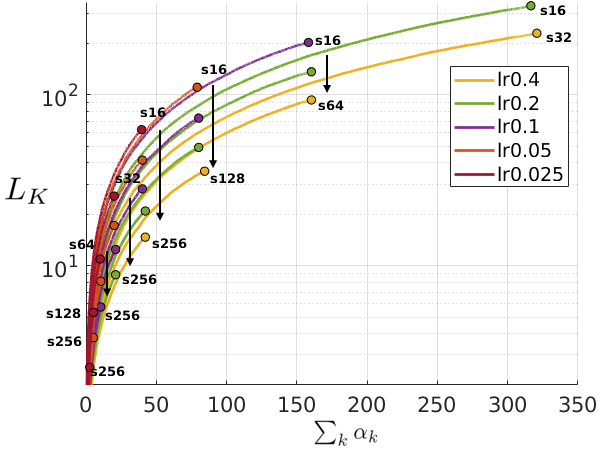} \\
(c) $\bar C_K$ on the training  &
(d) $L_K$ on the training \\
\end{tabular}}
\end{center}
\caption{The re-plot of Fig.7-(d,e) from the main article, , showing $\bar C_k$ and $L_k$ as a function of the cumulative learning rate, i.e., $\sum_k \alpha_k$ during the training.}
\label{fig:m2lp_iterlr}
\end{figure*}

Fig.~\ref{fig:intro_supp} is a new version of Fig.~1 (from the main manuscript), including the time scale represented by the first 320 iterations $k \in \{1,...,320\}$ (where $k$ represents the number of sampled iterations),
and $\sum_k \alpha_k$, where $\alpha_k$ represents the learning rate at the $k^{th}$ iteration.
This is equivalent to \{160, 80, 40, 20, 10, 5, 2.5\} training epochs with mini batches of size in \{512, 256, 128, 64, 32, 16, 8\}, respectively.
Fig.~\ref{fig:cond_eig_supp} is a new version of Fig.~2 (from the main manuscript), showing the recorded measures of $\bar C_K$ and $L_K$ for the full training of 320 epochs.
Figures~\ref{fig:intro_supp} and~\ref{fig:cond_eig_supp} can be used to reach the same conclusion reached in the main manuscript, which is that the proposed measures can be used monitor the training, where the relative difference between the measured values stay consistently sorted over the full course of the training, and high values for either measure indicate poor training convergence or generalisation.
Fig.~\ref{fig:functional_relationship_supp} and Fig.~\ref{fig:m2lp_iterlr} are the respective re-plots of Fig.~4 and Fig.~7-(d,e) (from the main manuscript), using the time scale $\sum_k \alpha_k$.
Finally, Fig.~\ref{fig:dynamic_sampling_runtime_iter} and Fig.~\ref{fig:dynamic_sampling_runtime_iterlr} are the re-plots of Fig.~6 (from the main manuscript) using the time scale $k$ and $\sum_k \alpha_k$ respectively.

\section{Training and testing results}
In this supplementary material, we also show the training and testing results for the four datasets used to plot the graphs in this work.
The top five most accurate models of each dataset in terms of mini-batch size and learning rate are highlighted.
The most accurate model of the dynamic sampling method is marked separately.

\FloatBarrier

\begin{table*}[!htbp]
\begin{center}

\resizebox{\textwidth}{!}{
\begin{tabular}{|l|c|c|c|c|c|c|c|}
\hline
\multicolumn{8}{|c|}{} \\
\multicolumn{8}{|c|}{ResNet CIFAR-10 training error} \\
\multicolumn{8}{|c|}{} \\
\hline
\multicolumn{8}{|c|}{Mini-batch size and learning rate} \\
\hline
$\alpha_k$ \textbackslash $ \text{   } {|\mathcal{B}_k|}$ & 8 & 16 & 32 & 64 & 128 & 256 & 512 \\
\hline
0.025 &  $0.03\pm0.02$ & $0.01\pm0.00$ & $0.00\pm0.00$ & $0.00\pm0.00$ & $0.00\pm0.00$ & $0.01\pm0.00$ & $0.01\pm0.00$ \\
\hline
0.05  & $0.05\pm0.00$ & $0.02\pm0.00$ & $0.01\pm0.00$ & $0.00\pm0.00$ & $0.00\pm0.00$ & $0.00\pm0.00$ & $0.01\pm0.00$ \\
\hline
0.1   & $0.55\pm0.05$ & $0.03\pm0.01$ & $0.01\pm0.01$ & $0.00\pm0.00$ & $0.00\pm0.00$ & $0.00\pm0.00$ & $0.00\pm0.00$ \\
\hline
0.2   & $3.05\pm0.37$ & $0.49\pm0.03$ & $0.02\pm0.01$ & $0.01\pm0.00$ & $0.00\pm0.00$ & $0.00\pm0.00$ & $0.00\pm0.00$ \\
\hline
0.4   & $15.75\pm1.38$ & $4.04\pm0.28$ & $0.70\pm0.10$ & $0.03\pm0.01$ & $0.01\pm0.01$ & $0.00\pm0.00$ & $0.00\pm0.00$ \\
\hline
\multicolumn{8}{|c|}{Dynamic Sampling Alternatives}\\
\hline
& s16-to-256  & s32-to-128 & s32-to-512 & s128-to-512 & s512-to-32 & & \\
\hline
0.1   & $0.00\pm0.00$ & $0.00\pm0.00$ & $0.00\pm0.00$ & $0.00\pm0.00$ & $0.02\pm0.01$ & & \\
\hline
& s16-to-256-MS & s32-to-128-MS & s32-to-512-MS & s128-to-512-MS & & & \\
\hline
0.1 & $0.00\pm0.00$ & $0.00\pm0.00$ & $0.00\pm0.00$ & $0.00\pm0.00$ & & &\\
\hline

\multicolumn{8}{|c|}{} \\
\multicolumn{8}{|c|}{ResNet CIFAR-10 testing error} \\
\multicolumn{8}{|c|}{} \\
\hline
\multicolumn{8}{|c|}{Mini-batch size and learning rate} \\
\hline
$\alpha_k$ \textbackslash $ \text{   } {|\mathcal{B}_k|}$ & 8 & 16 & 32 & 64 & 128 & 256 & 512 \\
\hline
0.025 &  $\bf 4.77 \pm 0.04$ & $\bf 4.76 \pm 0.11$ & $4.98 \pm 0.08$ & $5.81 \pm 0.19$ & $6.52 \pm 0.06$ & $7.41 \pm 0.12$ & $8.64 \pm 0.14$ \\
\hline
0.05  & $5.09\pm0.04$ & $\bf 4.80\pm0.10$ & $\bf 4.67\pm0.22$ & $5.00\pm0.09$ & $5.74\pm0.03$ & $6.64\pm0.15$ & $7.50\pm0.09$ \\
\hline
0.1   & $6.24\pm0.17$ & $5.11\pm0.21$ & $\bf 4.78\pm0.05$ & $4.95\pm0.09$ & $5.11\pm0.14$ & $5.65\pm0.12$ & $6.55\pm0.16$ \\
\hline
0.2   & $8.45\pm0.41$ & $6.47\pm0.38$ & $5.29\pm0.31$ & $4.98\pm0.12$ & $4.94\pm0.15$ & $5.15\pm0.24$ & $5.72\pm0.06$ \\
\hline
0.4   & $17.23\pm0.58$ & $9.42\pm0.69$ & $7.10\pm0.30$ & $5.55\pm0.12$ & $4.91\pm0.17$ & $4.95\pm0.11$ & $5.38\pm0.02$ \\
\hline
\multicolumn{8}{|c|}{Dynamic Sampling Alternatives}\\
\hline
& s16-to-256  & s32-to-128 & s32-to-512 & s128-to-512 & s512-to-32 & & \\
\hline
0.1   & $5.07\pm0.13$ & $\bf 4.90\pm0.05$ & $5.07\pm0.21$ & $5.33\pm0.19$ & $5.56\pm0.09$ & & \\
\hline
& s16-to-256-MS & s32-to-128-MS & s32-to-512-MS & s128-to-512-MS & & & \\
\hline
0.1 & $\bf 4.76\pm0.22$ & $\bf 4.76\pm0.13$ & $4.97\pm0.13$ & $5.29\pm0.15$ & & &\\
\hline
\end{tabular}}
\end{center}
\end{table*}

\begin{table*}[!htbp]
\begin{center}
\resizebox{0.9\textwidth}{!}{
\begin{tabular}{|l|c|c|c|c|c|c|}
\hline
\multicolumn{7}{|c|}{} \\
\multicolumn{7}{|c|}{DenseNet CIFAR-10 training error} \\
\multicolumn{7}{|c|}{} \\
\hline
\multicolumn{7}{|c|}{Mini-batch size and learning rate} \\
\hline
$\alpha_k$ \textbackslash $ \text{   } {|\mathcal{B}_k|}$ & 8 & 16 & 32 & 64 & 128 & 256\\
\hline
0.025 &  $0.09\pm0.01$ & $0.01\pm0.01$ & $0.02\pm0.01$ & $0.01\pm0.01$ & $0.01\pm0.00$ & $0.01\pm0.00$  \\
\hline
0.05  & $0.22\pm0.01$ & $0.04\pm0.01$ & $0.01\pm0.01$ & $0.01\pm0.01$ & $0.00\pm0.00$ & $0.01\pm0.01$  \\
\hline
0.1   & $1.12\pm0.14$ & $0.12\pm0.01$ & $0.03\pm0.01$ & $0.00\pm0.00$ & $0.01\pm0.00$ & $0.00\pm0.00$  \\
\hline
0.2   & $5.18\pm0.44$ & $1.02\pm0.09$ & $0.10\pm0.00$ & $0.02\pm0.01$ & $0.00\pm0.00$ & $0.00\pm0.00$  \\
\hline
0.4   & $81.07\pm15.60$ & $5.85\pm0.29$ & $0.93\pm0.06$ & $0.07\pm0.01$ & $0.01\pm0.01$ & $0.00\pm0.00$  \\
\hline
\multicolumn{7}{|c|}{Dynamic Sampling Alternatives}\\
\hline
& s16-to-256  & s32-to-128 & s64-to-256 & s256-to-16 & & \\
\hline
0.1   & $0.00\pm0.00$ & $0.00\pm0.00$ & $0.00\pm0.00$ & $0.13\pm0.02$ & & \\
\hline
& s16-to-256-MS & s32-to-128-MS & s64-to-256-MS & 
& & \\
\hline
0.1 & $0.00\pm0.00$ & $0.00\pm0.00$ & $0.00\pm0.00$ & 
& & \\
\hline

\multicolumn{7}{|c|}{} \\
\multicolumn{7}{|c|}{DenseNet CIFAR-10 testing error} \\
\multicolumn{7}{|c|}{} \\
\hline
\multicolumn{7}{|c|}{Mini-batch size and learning rate} \\
\hline
$\alpha_k$ \textbackslash $ \text{   } {|\mathcal{B}_k|}$ & 8 & 16 & 32 & 64 & 128 & 256\\
\hline
0.025 &  $\bf 4.99\pm0.17$ & $\bf 4.82\pm0.07$ & $5.22\pm0.23$ & $6.04\pm0.08$ & $7.31\pm0.02$ & $8.93\pm0.39$  \\
\hline
0.05  & $5.67\pm0.10$ & $\bf 4.95\pm0.09$ & $\bf 4.82\pm0.29$ & $5.24\pm0.14$ & $6.08\pm0.14$ & $7.38\pm0.01$ \\
\hline
0.1   & $6.80\pm0.43$ & $5.48\pm0.09$ & $\bf 4.96\pm0.12$ & $5.00\pm0.20$ & $5.38\pm0.12$ & $6.04\pm0.35$  \\
\hline
0.2   & $9.36\pm0.17$ & $7.21\pm0.34$ & $5.53\pm0.08$ & $5.28\pm0.12$ & $5.06\pm0.18$ & $5.53\pm0.21$  \\
\hline
0.4   & $79.68\pm17.87$ & $10.42\pm0.20$ & $7.56\pm0.34$ & $5.76\pm0.22$ & $5.25\pm0.24$ & $5.29\pm0.13$  \\
\hline
\multicolumn{7}{|c|}{Dynamic Sampling Alternatives}\\
\hline
& s16-to-256  & s32-to-128 & s64-to-256 & s256-to-16 & & \\
\hline
0.1   & $5.60\pm0.21$ & $5.15\pm0.05$ & $\bf 5.03\pm0.03$ & $5.34\pm0.14$ & & \\
\hline
& s16-to-256-MS & s32-to-128-MS & s64-to-256-MS & 
& & \\
\hline
0.1 & $5.14\pm0.16$ & $\bf 4.63\pm0.10$ & $5.03\pm0.21$ & 
& & \\
\hline
\end{tabular}}
\end{center}
\end{table*}


\begin{table*}[!htbp]
\begin{center}

\resizebox{\textwidth}{!}{
\begin{tabular}{|l|c|c|c|c|c|c|c|}
\hline
\multicolumn{8}{|c|}{} \\
\multicolumn{8}{|c|}{ResNet CIFAR-100 training error} \\
\multicolumn{8}{|c|}{} \\
\hline
\multicolumn{8}{|c|}{Mini-batch size and learning rate} \\
\hline
$\alpha_k$ \textbackslash $ \text{   } {|\mathcal{B}_k|}$ & 8 & 16 & 32 & 64 & 128 & 256 & 512 \\
\hline
0.025 &  $0.56\pm0.04$ & $0.19\pm0.03$ & $0.09\pm0.01$ & $0.05\pm0.01$ & $0.03\pm0.00$ & $0.03\pm0.01$ & $0.05\pm0.01$ \\
\hline
0.05  & $1.16\pm0.01$ & $0.22\pm0.01$ & $0.09\pm0.00$ & $0.05\pm0.01$ & $0.04\pm0.01$ & $0.03\pm0.00$ & $0.03\pm0.00$ \\
\hline
0.1   & $3.88\pm0.11$ & $0.53\pm0.02$ & $0.13\pm0.01$ & $0.05\pm0.01$ & $0.03\pm0.01$ & $0.03\pm0.00$ & $0.02\pm0.00$ \\
\hline
0.2   & $16.99\pm0.44$ & $3.16\pm0.37$ & $0.22\pm0.03$ & $0.08\pm0.01$ & $0.03\pm0.01$ & $0.02\pm0.00$ & $0.02\pm0.01$ \\
\hline
0.4   & $95.69\pm4.85$ & $23.02\pm1.71$ & $3.55\pm0.19$ & $0.16\pm0.02$ & $0.05\pm0.01$ & $0.03\pm0.00$ & $0.02\pm0.01$ \\
\hline
\multicolumn{8}{|c|}{Dynamic Sampling Alternatives}\\
\hline
& s16-to-256  & s32-to-128 & s32-to-512 & s128-to-512 & s512-to-32 & & \\
\hline
0.1   & $0.03\pm0.01$ & $0.03\pm0.01$ & $0.02\pm0.01$ & $0.02\pm0.00$ & $0.21\pm0.01$ & & \\
\hline
& s16-to-256-MS & s32-to-128-MS & s32-to-512-MS & s128-to-512-MS & & & \\
\hline
0.1 & $0.02\pm0.00$ & $0.02\pm0.00$ & $0.02\pm0.00$ & $0.02\pm0.00$ & & &\\
\hline

\multicolumn{8}{|c|}{} \\
\multicolumn{8}{|c|}{ResNet CIFAR-100 testing error} \\
\multicolumn{8}{|c|}{} \\
\hline
\multicolumn{8}{|c|}{Mini-batch size and learning rate} \\
\hline
$\alpha_k$ \textbackslash $ \text{   } {|\mathcal{B}_k|}$ & 8 & 16 & 32 & 64 & 128 & 256 & 512 \\
\hline
0.025 & $\bf 23.74\pm0.13$ & $\bf 23.77\pm0.27$ & $24.55\pm0.27$ & $25.50\pm0.24$ & $27.23\pm0.15$ & $29.79\pm0.12$ & $32.66\pm0.36$ \\
\hline
0.05  & $24.61\pm0.33$ & $\bf 23.39\pm0.13$ & $\bf 23.60\pm0.16$ & $24.79\pm0.17$ & $25.93\pm0.22$ & $27.64\pm0.19$ & $29.56\pm0.22$ \\
\hline
0.1   & $26.55\pm0.40$ & $24.24\pm0.29$ & $\bf 23.46\pm0.21$ & $24.34\pm0.08$ & $25.44\pm0.41$ & $26.25\pm0.05$ & $28.14\pm0.27$ \\
\hline
0.2   & $32.02\pm0.43$ & $26.91\pm0.10$ & $24.22\pm0.36$ & $24.04\pm0.03$ & $24.72\pm0.17$ & $25.28\pm0.30$ & $25.98\pm0.10$ \\
\hline
0.4   & $95.47\pm4.99$ & $34.32\pm0.96$ & $28.05\pm0.30$ & $25.30\pm0.31$ & $24.11\pm0.53$ & $24.56\pm0.07$ & $25.37\pm0.12$ \\
\hline
\multicolumn{8}{|c|}{Dynamic Sampling Alternatives}\\
\hline
& s16-to-256  & s32-to-128 & s32-to-512 & s128-to-512 & s512-to-32 & &\\
\hline
0.1   & $24.27\pm0.36$ & $\bf 23.90\pm0.31$ & $24.22\pm0.33$ & $25.05\pm0.03$ & $26.35\pm0.26$  & &\\
\hline
& s16-to-256-MS & s32-to-128-MS & s32-to-512-MS & s128-to-512-MS & & &\\
\hline
0.1 & $23.93\pm0.33$ & $\bf 23.69\pm0.34$ &  $24.14\pm0.20$ & $25.37\pm0.20$ & & & \\
\hline
\end{tabular}}
\end{center}
\end{table*}

\begin{table*}[!htbp]
\begin{center}
\resizebox{0.9\textwidth}{!}{
\begin{tabular}{|l|c|c|c|c|c|c|}
\hline
\multicolumn{7}{|c|}{} \\
\multicolumn{7}{|c|}{DenseNet CIFAR-100 training error} \\
\multicolumn{7}{|c|}{} \\
\hline
\multicolumn{7}{|c|}{Mini-batch size and learning rate} \\
\hline
$\alpha_k$ \textbackslash $ \text{   } {|\mathcal{B}_k|}$ & 8 & 16 & 32 & 64 & 128 & 256\\
\hline
0.025 &  $1.69\pm0.11$ & $0.34\pm0.03$ & $0.12\pm0.02$ & $0.06\pm0.01$ & $0.05\pm0.01$ & $0.09\pm0.01$  \\
\hline
0.05  & $4.41\pm0.23$ & $0.68\pm0.05$ & $0.13\pm0.01$ & $0.05\pm0.01$ & $0.04\pm0.00$ & $0.03\pm0.01$  \\
\hline
0.1   & $11.74\pm0.16$ & $2.30\pm0.15$ & $0.26\pm0.01$ & $0.06\pm0.00$ & $0.02\pm0.01$ & $0.03\pm0.00$  \\
\hline
0.2   & $26.97\pm1.54$ & $8.40\pm0.45$ & $1.00\pm0.06$ & $0.13\pm0.02$ & $0.04\pm0.00$ & $0.03\pm0.01$  \\
\hline
0.4   & $99.09\pm0.05$ & $33.54\pm3.69$ & $7.28\pm0.77$ & $0.61\pm0.08$ & $0.06\pm0.00$ & $0.03\pm0.00$  \\
\hline
\multicolumn{7}{|c|}{Dynamic Sampling Alternatives}\\
\hline
& s16-to-256  & s32-to-128 & s64-to-256 & s256-to-16 & & \\
\hline
0.1   & $0.08\pm0.01$ & $0.04\pm0.01$ & $0.03\pm0.00$ & $0.74\pm0.02$ & & \\
\hline
& s16-to-256-MS & s32-to-128-MS & s64-to-256-MS & 
& & \\
\hline
0.1 & $0.03\pm0.00$ & $0.02\pm0.00$ & $0.02\pm0.01$ & 
& & \\
\hline

\multicolumn{7}{|c|}{} \\
\multicolumn{7}{|c|}{DenseNet CIFAR-100 testing error} \\
\multicolumn{7}{|c|}{} \\
\hline
\multicolumn{7}{|c|}{Mini-batch size and learning rate} \\
\hline
$\alpha_k$ \textbackslash $ \text{   } {|\mathcal{B}_k|}$ & 8 & 16 & 32 & 64 & 128 & 256\\
\hline
0.025 &  $\bf 22.9\pm0.47$ & $\bf 23.29\pm0.23$ & $25.10\pm0.45$ & $26.61\pm0.05$ & $29.80\pm0.22$ & $33.71\pm0.17$  \\
\hline
0.05  & $24.44\pm0.44$ & $\bf 23.33\pm0.33$ & $\bf 23.64\pm0.23$ & $25.01\pm0.41$ & $26.97\pm0.11$ & $29.63\pm0.14$ \\
\hline
0.1   & $27.22\pm0.12$ & $24.63\pm0.73$ & $\bf 23.26\pm0.12$ & $24.19\pm0.30$ & $25.61\pm0.21$ & $27.45\pm0.33$  \\
\hline
0.2   & $33.19\pm1.04$ & $27.89\pm0.59$ & $25.04\pm0.24$ & $23.89\pm0.02$ & $24.38\pm0.35$ & $25.85\pm0.32$  \\
\hline
0.4   & $99.00\pm0.00$ & $38.82\pm2.43$ & $29.15\pm0.42$ & $25.84\pm0.43$ & $24.03\pm0.32$ & $24.64\pm0.37$  \\
\hline
\multicolumn{7}{|c|}{Dynamic Sampling Alternatives}\\
\hline
& s16-to-256  & s32-to-128 & s64-to-256 & s256-to-16 & & \\
\hline
0.1   & $24.71\pm0.90$ & $\bf 23.68\pm0.22$ & $24.20\pm0.29$ & $25.46\pm0.07$ & & \\
\hline
& s16-to-256-MS & s32-to-128-MS & s64-to-256-MS & 
& & \\
\hline
0.1 & $24.48\pm0.48$ & $\bf 23.92\pm0.11$ & $24.50\pm0.51$ & 
& & \\
\hline
\end{tabular}}
\end{center}
\end{table*}


\begin{table*}[!htbp]
\begin{center}

\resizebox{\textwidth}{!}{
\begin{tabular}{|l|c|c|c|c|c|c|c|}
\hline
\multicolumn{8}{|c|}{} \\
\multicolumn{8}{|c|}{ResNet SVHN training error} \\
\multicolumn{8}{|c|}{} \\
\hline
\multicolumn{8}{|c|}{Mini-batch size and learning rate} \\
\hline
$\alpha_k$ \textbackslash $ \text{   } {|\mathcal{B}_k|}$ & 8 & 16 & 32 & 64 & 128 & 256 & 512 \\
\hline
0.025 &  $0.21\pm0.01$ & $0.05\pm0.01$ & $0.01\pm0.00$ & $0.01\pm0.00$ & $0.01\pm0.00$ & $0.01\pm0.00$ & $0.03\pm0.00$ \\
\hline
0.05  & $0.46\pm0.01$ & $0.19\pm0.00$ & $0.04\pm0.00$ & $0.01\pm0.00$ & $0.01\pm0.00$ & $0.02\pm0.00$ & $0.02\pm0.00$ \\
\hline
0.1   & $0.89\pm0.01$ & $0.48\pm0.01$ & $0.19\pm0.01$ & $0.03\pm0.00$ & $0.01\pm0.00$ & $0.02\pm0.00$ & $0.02\pm0.00$ \\
\hline
0.2   & $1.58\pm0.04$ & $1.00\pm0.01$ & $0.51\pm0.00$ & $0.18\pm0.01$ & $0.03\pm0.00$ & $0.01\pm0.00$ & $0.02\pm0.00$ \\
\hline
0.4   & $82.73\pm0.00$ & $1.79\pm0.03$ & $1.09\pm0.00$ & $0.57\pm0.02$ & $0.20\pm0.00$ & $0.02\pm0.01$ & $0.02\pm0.00$ \\
\hline
\multicolumn{8}{|c|}{Dynamic Sampling Alternatives}\\
\hline
& s16-to-256  & s32-to-128 & s32-to-512 & s128-to-512 & s512-to-32 & & \\
\hline
0.1   & $0.05\pm0.00$ & $0.06\pm0.00$ & $0.02\pm0.00$ & $0.02\pm0.00$ & $0.22\pm0.01$ & & \\
\hline
& s16-to-256-MS & s32-to-128-MS & s32-to-512-MS & s128-to-512-MS & & & \\
\hline
0.1 & $0.00\pm0.00$ & $0.00\pm0.00$ & $0.00\pm0.00$ & $0.00\pm0.00$ & & &\\
\hline

\multicolumn{8}{|c|}{} \\
\multicolumn{8}{|c|}{ResNet SVHN testing error} \\
\multicolumn{8}{|c|}{} \\
\hline
\multicolumn{8}{|c|}{Mini-batch size and learning rate} \\
\hline
$\alpha_k$ \textbackslash $ \text{   } {|\mathcal{B}_k|}$ & 8 & 16 & 32 & 64 & 128 & 256 & 512 \\
\hline
0.025 & $2.04\pm0.14$ & $\bf 1.86\pm0.03$ & $\bf 1.93\pm0.01$ & $\bf 1.90\pm0.02$ & $1.98\pm0.02$ & $2.23\pm0.05$ & $2.53\pm0.09$ \\
\hline
0.05  & $2.09\pm0.07$ & $2.08\pm0.11$ & $1.98\pm0.02$ & $\bf 1.90\pm0.05$ & $1.97\pm0.01$ & $2.13\pm0.08$ & $2.28\pm0.03$ \\
\hline
0.1   & $2.18\pm0.03$ & $2.18\pm0.03$ & $2.05\pm0.06$ & $1.94\pm0.04$ & $\bf 1.93\pm0.04$ & $1.99\pm0.08$ & $2.13\pm0.10$ \\
\hline
0.2   & $2.89\pm0.05$ & $2.39\pm0.07$ & $2.07\pm0.05$ & $2.04\pm0.04$ & $2.01\pm0.03$ & $1.94\pm0.07$ & $2.07\pm0.04$ \\
\hline
0.4   & $80.41\pm0.00$ & $3.10\pm0.08$ & $2.49\pm0.07$ & $2.24\pm0.06$ & $2.10\pm0.09$ & $2.07\pm0.06$ & $1.99\pm0.06$ \\
\hline
\multicolumn{8}{|c|}{Dynamic Sampling Alternatives}\\
\hline
& s16-to-256  & s32-to-128 & s32-to-512 & s128-to-512 & s512-to-32 & &\\
\hline
0.1   & $2.06\pm0.06$ &  $1.96\pm0.04$ & $2.00\pm0.07$ & $\bf 1.91\pm0.01$ & $2.10\pm0.03$ & & \\
\hline
& s16-to-256-MS & s32-to-128-MS & s32-to-512-MS & s128-to-512-MS & & &\\
\hline
0.1 & $1.93\pm0.06$ & $1.96\pm0.04$ & $\bf 1.90\pm0.04$ & $1.93\pm0.02$ & & & \\
\hline
\end{tabular}}
\end{center}
\end{table*}

\begin{table*}[!htbp]
\begin{center}
\resizebox{0.9\textwidth}{!}{
\begin{tabular}{|l|c|c|c|c|c|c|}
\hline
\multicolumn{7}{|c|}{} \\
\multicolumn{7}{|c|}{DenseNet SVHN training error} \\
\multicolumn{7}{|c|}{} \\
\hline
\multicolumn{7}{|c|}{Mini-batch size and learning rate} \\
\hline
$\alpha_k$ \textbackslash $ \text{   } {|\mathcal{B}_k|}$ & 8 & 16 & 32 & 64 & 128 & 256\\
\hline
0.025 &  $0.36\pm0.02$ & $0.17\pm0.00$ & $0.09\pm0.00$ & $0.06\pm0.01$ & $0.07\pm0.00$ & $0.10\pm0.01$  \\
\hline
0.05  & $0.66\pm0.01$ & $0.34\pm0.00$ & $0.15\pm0.01$ & $0.07\pm0.01$ & $0.06\pm0.01$ & $0.07\pm0.00$  \\
\hline
0.1   & $1.15\pm0.00$ & $0.64\pm0.01$ & $0.30\pm0.01$ & $0.13\pm0.01$ & $0.07\pm0.01$ & $0.07\pm0.00$  \\
\hline
0.2   & $1.78\pm0.03$ & $1.15\pm0.05$ & $0.63\pm0.03$ & $0.28\pm0.00$ & $0.11\pm0.00$ & $0.10\pm0.01$  \\
\hline
0.4   & $82.73\pm0.00$ & $55.76\pm46.71$ & $28.36\pm47.08$ & $0.64\pm0.02$ & $0.30\pm0.01$ & $0.13\pm0.01$  \\
\hline
\multicolumn{7}{|c|}{Dynamic Sampling Alternatives}\\
\hline
& s16-to-256  & s32-to-128 & s64-to-256 & s256-to-16 & & \\
\hline
0.1   & $0.16\pm0.00$ & $0.18\pm0.01$ & $0.10\pm0.01$ & $0.53\pm0.00$ & & \\
\hline
& s16-to-256-MS & s32-to-128-MS & s64-to-256-MS & 
& & \\
\hline
0.1 & $0.05\pm0.01$ & $0.02\pm0.01$ & $0.01\pm0.00$ & 
& & \\
\hline

\multicolumn{7}{|c|}{} \\
\multicolumn{7}{|c|}{DenseNet SVHN testing error} \\
\multicolumn{7}{|c|}{} \\
\hline
\multicolumn{7}{|c|}{Mini-batch size and learning rate} \\
\hline
$\alpha_k$ \textbackslash $ \text{   } {|\mathcal{B}_k|}$ & 8 & 16 & 32 & 64 & 128 & 256\\
\hline
0.025 &  $2.04\pm0.02$ & $1.96\pm0.03$ & $\bf 1.89\pm0.01$ & $\bf 1.94\pm0.09$ & $2.14\pm0.06$ & $2.39\pm0.03$  \\
\hline
0.05  & $2.16\pm0.07$ & $2.08\pm0.02$ & $\bf 1.91\pm0.09$ & $\bf 1.93\pm0.03$ & $2.01\pm0.03$ & $2.09\pm0.05$  \\
\hline
0.1   & $2.63\pm0.13$ & $2.32\pm0.07$ & $2.16\pm0.05$ & $2.06\pm0.01$ & $\bf 1.93\pm0.09$ & $1.96\pm0.03$  \\
\hline
0.2   & $3.16\pm0.20$ & $2.59\pm0.15$ & $2.34\pm0.16$ & $2.14\pm0.19$ & $2.09\pm0.05$ & $2.06\pm0.04$  \\
\hline
0.4   & $80.41\pm0.00$ & $54.65\pm44.61$ & $28.56\pm44.91$ & $2.34\pm0.08$ & $2.22\pm0.12$ & $2.09\pm0.03$ \\
\hline
\multicolumn{7}{|c|}{Dynamic Sampling Alternatives}\\
\hline
& s16-to-256  & s32-to-128 & s64-to-256 & s256-to-16 & & \\
\hline
0.1   & $2.14\pm0.08$ & $2.03\pm0.07$ & $\bf 2.00\pm0.06$ & $2.07\pm0.10$ & & \\
\hline
& s16-to-256-MS & s32-to-128-MS & s64-to-256-MS & 
& & \\
\hline
0.1 & $2.11\pm0.04$ & $2.14\pm0.03$ & $\bf 2.03\pm0.02$ & 
& & \\
\hline
\end{tabular}}
\end{center}
\end{table*}


\begin{table*}[!htbp]
\begin{center}
\resizebox{\textwidth}{!}{
\begin{tabular}{|l|c|c|c|c|c|c|c|}
\hline
\multicolumn{7}{|c|}{} \\
\multicolumn{7}{|c|}{ResNet MNIST training error} \\
\multicolumn{7}{|c|}{} \\
\hline
\multicolumn{7}{|c|}{Mini-batch size and learning rate} \\
\hline
$\alpha_k$ \textbackslash $ \text{   } {|\mathcal{B}_k|}$ & 2 & 4 & 8 & 16 & 32 & 64 \\
\hline
0.00625 & $0.30\pm0.02$ & $0.33\pm0.02$ & $0.33\pm0.02$ & $0.36\pm0.02$ & $0.39\pm0.01$ & $0.48\pm0.01$ \\
\hline
0.0125 & $0.32\pm0.01$ & $0.35\pm0.01$ & $0.33\pm0.01$ & $0.31\pm0.01$ & $0.30\pm0.03$ & $0.32\pm0.02$ \\
\hline
0.025 & $0.34\pm0.02$ & $0.35\pm0.01$ & $0.34\pm0.01$ & $0.30\pm0.02$ & $0.27\pm0.01$ & $0.24\pm0.01$ \\
\hline
0.05  & $0.43\pm0.01$ & $0.39\pm0.03$ & $0.37\pm0.02$ & $0.30\pm0.02$ & $0.25\pm0.01$ & $0.22\pm0.02$  \\
\hline
0.1   & $0.51\pm0.01$ & $0.47\pm0.01$ & $0.39\pm0.01$ & $0.30\pm0.02$ & $0.27\pm0.01$ & $0.20\pm0.01$  \\
\hline
0.2   & $0.89\pm0.12$ & $0.57\pm0.02$ & $0.46\pm0.01$ & $0.36\pm0.02$ & $0.28\pm0.00$ & $0.25\pm0.01$  \\
\hline
0.4   & $89.53\pm0.00$ & $1.15\pm0.18$ & $0.57\pm0.01$ & $0.42\pm0.04$ & $0.30\pm0.01$ & $0.23\pm0.01$  \\
\hline
0.8   & $89.53\pm0.00$ & $2.39\pm0.34$ & $0.60\pm0.01$ & $0.39\pm0.02$ & $0.28\pm0.02$ & $0.24\pm0.01$  \\
\hline
& 128 & 256 & 512 & 1024 & 2048 & \\
\hline
0.00625 & $0.66\pm0.01$ & $1.03\pm0.03$ & $1.69\pm0.01$ & $2.98\pm0.00$ & $7.11\pm0.12$ & \\
\hline
0.0125 & $0.44\pm0.03$ & $0.62\pm0.01$ & $1.04\pm0.02$ & $1.69\pm0.03$ & $3.09\pm0.06$ & \\
\hline
0.025 & $0.28\pm0.01$ & $0.40\pm0.01$ & $0.62\pm0.02$ & $1.02\pm0.01$ & $1.78\pm0.02$ & \\
\hline
0.05  & $0.22\pm0.02$ & $0.27\pm0.02$ & $0.39\pm0.01$ & $0.62\pm0.01$ & $1.07\pm0.03$ & \\
\hline
0.1   & $0.19\pm0.02$ & $0.21\pm0.01$ & $0.26\pm0.01$ & $0.38\pm0.01$ & $0.63\pm0.03$ & \\
\hline
0.2   & $0.20\pm0.01$ & $0.18\pm0.01$ & $0.20\pm0.01$ & $0.25\pm0.02$ & $0.45\pm0.02$ & \\
\hline
0.4   & $0.22\pm0.02$ & $0.17\pm0.01$ & $0.20\pm0.01$ & $0.22\pm0.02$ & $0.33\pm0.03$ & \\
\hline
0.8   & $0.20\pm0.01$ & $0.19\pm0.01$ & $0.17\pm0.02$ & $0.37\pm0.12$ & $89.82\pm0.10$ & \\
\hline
\multicolumn{7}{|c|}{Dynamic Sampling Alternatives}\\
\hline
& s16-to-64 & s16-to-256  & s32-to-128 & s32-to-512 & s128-to-512 & s512-to-32 \\
\hline
0.1  & $0.22\pm0.01$ & $0.20\pm0.00$ & $0.19\pm0.02$ &  $0.16\pm0.00$ & $0.17\pm0.01$ & $0.42\pm0.02$ \\
\hline
& s16-to-64-MS & s16-to-256-MS & s32-to-128-MS & s32-to-512-MS & s128-to-512-MS & \\
\hline
0.1  & $0.15\pm0.00$ & $0.15\pm0.01$ & $0.13\pm0.01$ & $0.12\pm0.02$ & $0.15\pm0.01$ & \\
\hline

\multicolumn{7}{|c|}{} \\
\multicolumn{7}{|c|}{MNIST testing error} \\
\multicolumn{7}{|c|}{} \\
\hline
\multicolumn{7}{|c|}{Mini-batch size and learning rate} \\
\hline
$\alpha_k$ \textbackslash $ \text{   } {|\mathcal{B}_k|}$ & 2 & 4 & 8 & 16 & 32 & 64 \\
\hline
0.00625 & $0.46\pm0.03$ & $0.42\pm0.03$ & $0.40\pm0.04$ & $0.44\pm0.01$ & $0.50\pm0.03$ & $0.56\pm0.02$ \\
\hline
0.0125 & $0.41\pm0.01$ & $\bf 0.38\pm0.04$ & $\bf 0.37\pm0.05$ & $0.39\pm0.00$ & $0.46\pm0.03$ & $0.49\pm0.05$ \\
\hline
0.025 & $0.48\pm0.04$ & $0.40\pm0.01$ & $\bf 0.37\pm0.03$ & $0.38\pm0.05$ & $0.42\pm0.06$ & $0.46\pm0.02$ \\
\hline
0.05  & $0.43\pm0.06$ & $0.40\pm0.06$ & $0.39\pm0.02$ & $\bf 0.38\pm0.02$ & $0.39\pm0.03$ & $0.39\pm0.03$  \\
\hline
0.1   & $0.49\pm0.03$ & $0.44\pm0.07$ & $0.39\pm0.02$ & $\bf 0.36\pm0.02$ & $0.39\pm0.03$ & $0.38\pm0.06$  \\
\hline
0.2   & $0.76\pm0.23$ & $0.45\pm0.06$ & $0.44\pm0.03$ & $0.43\pm0.05$ & $0.40\pm0.04$ & $0.41\pm0.05$  \\
\hline
0.4   & $89.90\pm0.00$ & $0.84\pm0.20$ & $0.49\pm0.06$ & $0.42\pm0.01$ & $0.42\pm0.02$ & $0.39\pm0.04$  \\
\hline
0.8   & $89.90\pm0.00$ & $89.90\pm0.00$ & $1.92\pm0.49$ & $0.52\pm0.04$ & $0.51\pm0.04$ & $0.44\pm0.02$  \\
\hline
& 128 & 256 & 512 & 1024 & 2048 \\
\hline
0.00625 & $0.70\pm0.02$ & $1.02\pm0.03$ & $1.60\pm0.05$ & $2.78\pm0.05$ & $6.28\pm0.01$ & \\
\hline
0.0125 & $0.60\pm0.04$ & $0.75\pm0.02$ & $1.02\pm0.04$ & $1.62\pm0.01$ & $2.80\pm0.02$ & \\
\hline
0.025 & $0.48\pm0.02$ & $0.60\pm0.06$ & $0.77\pm0.01$ & $0.96\pm0.01$ & $1.61\pm0.03$ & \\
\hline
0.05  & $0.44\pm0.01$ & $0.50\pm0.05$ & $0.65\pm0.03$ & $0.80\pm0.03$ & $1.12\pm0.02$ &  \\
\hline
0.1   & $0.40\pm0.06$ & $0.47\pm0.05$ & $0.57\pm0.03$ & $0.69\pm0.04$ & $0.87\pm0.05$ &  \\
\hline
0.2   & $0.40\pm0.03$ & $0.40\pm0.03$ & $0.47\pm0.08$ & $0.63\pm0.01$ & $0.70\pm0.02$ &  \\
\hline
0.4   & $0.40\pm0.07$ & $0.42\pm0.02$ & $0.44\pm0.05$ & $0.53\pm0.05$ & $0.65\pm0.02$ &  \\
\hline
0.8   & $0.41\pm0.04$ & $0.39\pm0.06$ & $0.40\pm0.01$ & $0.47\pm0.02$ & $0.62\pm0.14$ &  \\
\hline
\multicolumn{7}{|c|}{Dynamic Sampling Alternatives}\\
\hline
& s16-to-64 & s16-to-256  & s32-to-128 & s32-to-512 & s128-to-512 & s512-to-32 \\
\hline
0.1  & $\bf 0.39\pm0.02$ & $0.41\pm0.04$ & $\bf 0.39\pm0.04$ &  $\bf 0.39\pm0.05$ & $0.40\pm0.04$ & $0.45\pm0.05$ \\
\hline
& s16-to-64-MS & s16-to-256-MS & s32-to-128-MS & s32-to-512-MS & s128-to-512-MS & \\
\hline
0.1  & $\bf 0.34\pm0.02$ & $0.40\pm0.03$ & $0.41\pm0.02$ & $0.43\pm0.05$ & $0.42\pm0.01$ & \\
\hline
\end{tabular}}
\end{center}
\end{table*}

\begin{table*}[!htbp]
\begin{center}
\resizebox{\textwidth}{!}{
\begin{tabular}{|l|c|c|c|c|c|c|}
\hline
\multicolumn{7}{|c|}{} \\
\multicolumn{7}{|c|}{DenseNet MNIST training error} \\
\multicolumn{7}{|c|}{} \\
\hline
\multicolumn{7}{|c|}{Mini-batch size and learning rate} \\
\hline
$\alpha_k$ \textbackslash $ \text{   } {|\mathcal{B}_k|}$ & 2 & 4 & 8 & 16 & 32 & 64 \\
\hline
0.00625 & $0.80\pm0.02$ & $0.82\pm0.02$ & $0.88\pm0.02$ & $0.87\pm0.01$ & $1.01\pm0.01$ & $1.23\pm0.02$   \\
\hline
0.0125 & $0.78\pm0.07$ & $0.82\pm0.02$ & $0.78\pm0.04$ & $0.75\pm0.03$ & $0.75\pm0.00$ & $0.88\pm0.02$  \\
\hline
0.025 & $0.86\pm0.04$ & $0.81\pm0.03$ & $0.73\pm0.03$ & $0.70\pm0.02$ & $0.69\pm0.03$ & $0.72\pm0.03$  \\
\hline
0.05  & $0.88\pm0.03$ & $0.83\pm0.01$ & $0.74\pm0.04$ & $0.67\pm0.04$ & $0.62\pm0.02$ & $0.56\pm0.03$  \\
\hline
0.1   & $1.13\pm0.09$ & $0.92\pm0.05$ & $0.79\pm0.02$ & $0.67\pm0.04$ & $0.59\pm0.04$ & $0.54\pm0.03$  \\
\hline
0.2   & $4.54\pm1.19$ & $1.18\pm0.10$ & $0.93\pm0.01$ & $0.71\pm0.02$ & $0.61\pm0.02$ & $0.56\pm0.00$  \\
\hline
0.4   & $89.38\pm0.00$ & $89.32\pm0.00$ & $1.19\pm0.12$ & $0.82\pm0.01$ & $0.63\pm0.01$ & $0.55\pm0.02$ \\
\hline
0.8   & $89.63\pm0.00$ &$89.34\pm0.00$ & $32.85\pm48.96$ & $1.05\pm0.11$ & $0.72\pm0.00$ & $0.57\pm0.01$ \\
\hline
$\alpha_k$ \textbackslash $\text{   } {|\mathcal{B}_k|}$ & 128 & 256 & 512 & 1024 & 2048 & \\
\hline
0.00625 & $1.67\pm0.05$ & $2.52\pm0.03$ & $4.36\pm0.02$ & $10.03\pm0.11$ & $28.40\pm0.05$ &  \\
\hline
0.0125 & $1.17\pm0.03$ & $1.57\pm0.01$ & $2.49\pm0.02$ & $4.33\pm0.07$ & $10.22\pm0.08$ &  \\
\hline
0.025 & $0.83\pm0.03$ & $1.09\pm0.03$ & $1.62\pm0.03$ & $2.47\pm0.01$ & $4.23\pm0.04$ &  \\
\hline
0.05  & $0.65\pm0.01$ & $0.81\pm0.01$ & $1.08\pm0.05$ & $1.59\pm0.03$ & $2.40\pm0.07$ &  \\
\hline
0.1   & $0.52\pm0.03$ & $0.61\pm0.02$ & $0.80\pm0.01$ & $1.13\pm0.02$ & $1.67\pm0.02$ &  \\
\hline
0.2   & $0.47\pm0.03$ & $0.48\pm0.01$ & $0.59\pm0.03$ & $0.79\pm0.01$ & $1.17\pm0.01$ &  \\
\hline
0.4   & $0.50\pm0.00$ & $0.45\pm0.00$ & $0.48\pm0.01$ & $0.59\pm0.05$ & $0.85\pm0.03$ &  \\
\hline
0.8   & $0.50\pm0.01$ & $0.44\pm0.02$ & $0.46\pm0.01$ & $0.53\pm0.05$ & $0.67\pm0.04$ &  \\
\hline
\multicolumn{7}{|c|}{Dynamic Sampling Alternatives}\\
\hline
& s16-to-64 & s16-to-256  & s32-to-128 & s32-to-512 & s128-to-512 & s512-to-32 \\
\hline
0.1  & $0.49\pm0.01$ & $0.45\pm0.02$ & $0.45\pm0.02$ &  $0.43\pm0.02$ & $0.50\pm0.02$ & $0.88\pm0.05$ \\
\hline
& s16-to-64-MS & s16-to-256-MS & s32-to-128-MS & s32-to-512-MS & s128-to-512-MS & \\
\hline
0.1  & $0.44\pm0.01$ & $0.41\pm0.03$ & $0.44\pm0.03$ & $0.44\pm0.01$ & $0.53\pm0.02$ & \\
\hline

\multicolumn{7}{|c|}{} \\
\multicolumn{7}{|c|}{DenseNet MNIST testing error} \\
\multicolumn{7}{|c|}{} \\
\hline
\multicolumn{7}{|c|}{Mini-batch size and learning rate} \\
\hline
$\alpha_k$ \textbackslash $ \text{   } {|\mathcal{B}_k|}$ & 2 & 4 & 8 & 16 & 32 & 64 \\
\hline
0.00625 & $0.94\pm0.14$ & $0.67\pm0.02$ & $0.73\pm0.07$ & $0.76\pm0.01$ & $0.90\pm0.05$ & $1.23\pm0.05$  \\
\hline
0.0125 & $0.78\pm0.10$ & $0.71\pm0.05$ & $0.66\pm0.04$ & $0.65\pm0.14$ & $0.76\pm0.03$ & $0.89\pm0.02$  \\
\hline
0.025 & $0.82\pm0.05$ & $0.66\pm0.08$ & $0.62\pm0.09 $ & $0.66\pm0.02$ & $0.70\pm0.08$ & $0.81\pm0.03$  \\
\hline
0.05  & $0.81\pm0.07$ & $0.63\pm0.02$ & $0.62\pm0.02$ & $\bf 0.61\pm0.02$ & $0.67\pm0.07$ & $0.73\pm0.01$  \\
\hline
0.1   & $0.93\pm0.05$ & $0.71\pm0.09$ & $0.61\pm0.06$ & $0.62\pm0.06$ & $ 0.62\pm0.06$ & $0.63\pm0.08$  \\
\hline
0.2   & $4.97\pm1.52$ & $0.84\pm0.09$ & $0.69\pm0.03$ & $\bf 0.60\pm0.04$ & $\bf 0.59\pm0.05$ & $0.66\pm0.03$  \\
\hline
0.4   & $88.65\pm0.00$ & $88.65\pm0.00$ & $0.77\pm0.09$ & $0.66\pm0.06$ & $\bf 0.54\pm0.02$ & $\bf 0.59\pm0.03$  \\
\hline
0.8   & $90.20\pm0.00$ & $88.65\pm0.00$ & $31.96\pm49.11$ & $0.76\pm0.11$ & $0.62\pm0.05$ & $0.65\pm0.03$ \\
\hline
$\alpha_k$ \textbackslash $\text{   } {|\mathcal{B}_k|}$ & 128 & 256 & 512 & 1024 & 2048 & \\
\hline
0.00625 & $1.68\pm0.05$ & $2.36\pm0.06$ & $4.08\pm0.07$ & $9.54\pm0.19$ & $29.71\pm0.09$ &  \\
\hline
0.0125 & $1.22\pm0.09$ & $1.60\pm0.08$ & $2.38\pm0.04$ & $3.97\pm0.06$ & $9.23\pm0.09$ &  \\
\hline
0.025 & $0.97\pm0.07$ & $1.21\pm0.05$ & $1.71\pm0.01$ & $2.51\pm0.04$ & $3.90\pm0.05$ &  \\
\hline
0.05  & $0.80\pm0.01$ & $1.02\pm0.01$ & $1.26\pm0.04$ & $1.65\pm0.09$ & $2.47\pm0.03$ &  \\
\hline
0.1   & $0.71\pm0.09$ & $0.77\pm0.06$ & $1.01\pm0.05$ & $1.17\pm0.08$ & $1.76\pm0.06$ &  \\
\hline
0.2   & $0.65\pm0.03$ & $0.66\pm0.05$ & $0.85\pm0.12$ & $0.99\pm0.06$ & $1.40\pm0.03$ &  \\
\hline
0.4   & $0.66\pm0.02$ & $0.64\pm0.01$ & $0.72\pm0.04$ & $0.82\pm0.03$ & $1.02\pm0.03$ &  \\
\hline
0.8   & $0.70\pm0.03$ & $0.67\pm0.08$ & $0.68\pm0.04$ & $0.71\pm0.08$ & $0.88\pm0.02$ &  \\
\hline
\multicolumn{7}{|c|}{Dynamic Sampling Alternatives}\\
\hline
& s16-to-64 & s16-to-256  & s32-to-128 & s32-to-512 & s128-to-512 & s512-to-32 \\
\hline
0.1  & $0.62\pm0.08$ & $0.59\pm0.05$ & $\bf 0.57\pm0.03$ &  $0.59\pm0.03$ & $0.71\pm0.04$ & $0.77\pm0.06$ \\
\hline
& s16-to-64-MS & s16-to-256-MS & s32-to-128-MS & s32-to-512-MS & s128-to-512-MS & \\
\hline
0.1  & $\bf 0.60\pm0.01$ & $0.60\pm0.03$ & $0.63\pm0.03$ &  $0.65\pm0.02$ & $0.82\pm0.08$ &  \\
\hline
\end{tabular}}
\end{center}
\end{table*}

\begin{table*}[!htbp]
\begin{center}
\resizebox{\textwidth}{!}{
\begin{tabular}{|l|c|c|c|c|c|c|}
\hline
\multicolumn{7}{|c|}{} \\
\multicolumn{7}{|c|}{2-layer MLP MNIST training error} \\
\multicolumn{7}{|c|}{} \\
\hline
\multicolumn{7}{|c|}{Mini-batch size and learning rate} \\
\hline
$\alpha_k$ \textbackslash $ \text{   } {|\mathcal{B}_k|}$ & 2 & 4 & 8 & 16 & 32 & 64 \\
\hline
0.00625 & $88.65+0.00$ & $2.82+0.09$ & $1.29+0.02$ & $0.91+0.06$ & $0.95+0.05$ & $1.01+0.04$   \\
\hline
0.0125 & $88.65+0.00$ & $3.54+0.06$ & $1.27+0.09$ & $1.00+0.09$ & $0.88+0.10$ & $0.93+0.12$  \\
\hline
0.025 & $88.65+0.00$ & $4.97+0.26$ & $1.39+0.08$ & $0.93+0.06$ & $0.86+0.04$ & $0.91+0.03$  \\
\hline
0.05  & $88.65+0.00$ & $88.65+0.00$ & $2.03+0.10$ & $1.02+0.05$ & $0.87+0.06$ & $0.82+0.02$  \\
\hline
0.1   & $88.65+0.00$ & $88.65+0.00$ & $3.23+0.27$ & $1.35+0.05$ & $0.87+0.03$ & $0.90+0.00$  \\
\hline
0.2   & $88.65+0.00$ & $88.65+0.00$ & $88.65+0.00$ & $2.12+0.11$ & $1.00+0.04$ & $0.93+0.08$  \\
\hline
0.4   & $88.65+0.00$ & $88.65+0.00$ & $88.65+0.00$ & $88.65+0.00$ & $1.86+0.05$ & $1.04+0.05$ \\
\hline
0.8   & $90.20+0.00$ &$88.65+0.00$ & $88.65+0.00$ & $88.65+0.00$ & $88.65+0.00$ & $1.60+0.09$ \\
\hline
$\alpha_k$ \textbackslash $\text{   } {|\mathcal{B}_k|}$ & 128 & 256 & 512 & 1024 &  & \\
\hline
0.00625 & $1.14+0.02$ & $1.35+0.01$ & $1.56+0.03$ & $2.02+0.07$ & &  \\
\hline
0.0125 & $1.11+0.06$ & $1.17+0.09$ & $1.36+0.02$ & $1.68+0.03$ & &  \\
\hline
0.025 & $0.95+0.09$ & $1.07+0.08$ & $1.15+0.06$ & $1.40+0.05$ & &  \\
\hline
0.05  & $0.90+0.06$ & $1.01+0.09$ & $1.07+0.02$ & $1.22+0.05$ & &  \\
\hline
0.1   & $0.90+0.02$ & $0.92+0.07$ & $1.06+0.03$ & $1.13+0.02$ & &  \\
\hline
0.2   & $0.90+0.05$ & $0.96+0.08$ & $0.89+0.12$ & $1.11+0.09$ & &  \\
\hline
0.4   & $0.77+0.04$ & $0.92+0.07$ & $0.89+0.08$ & $0.92+0.06$ & &  \\
\hline
0.8   & $0.96+0.07$ & $0.90+0.10$ & $0.93+0.02$ & $0.98+0.05$ & &  \\
\hline
\multicolumn{7}{|c|}{Dynamic Sampling Alternatives}\\
\hline
& s16-to-64 & s16-to-256  & s32-to-128 & s32-to-512 & s128-to-512 & s512-to-32 \\
\hline
0.1  & $1.07+0.06$ & $0.98+0.06$ & $0.79+0.10$ &  $0.93+0.02$ & $0.88+0.02$ & $0.95+0.05$ \\
\hline
& s16-to-64-MS & s16-to-256-MS & s32-to-128-MS & s32-to-512-MS & s128-to-512-MS & \\
\hline
0.1  & $0.90+0.12$ & $0.93+0.08$ & $0.73+0.08$ & $1.23+0.10$ & $0.91+0.04$ & \\
\hline

\multicolumn{7}{|c|}{} \\
\multicolumn{7}{|c|}{2-layer MLP MNIST testing error} \\
\multicolumn{7}{|c|}{} \\
\hline
\multicolumn{7}{|c|}{Mini-batch size and learning rate} \\
\hline
$\alpha_k$ \textbackslash $ \text{   } {|\mathcal{B}_k|}$ & 2 & 4 & 8 & 16 & 32 & 64 \\
\hline
0.00625 & $88.65+0.00$ & $2.82+0.09$ & $1.29+0.02$ & $0.91+0.06$ & $0.95+0.05$ & $1.01+0.04$  \\
\hline
0.0125 & $88.65+0.00$ & $3.54+0.06$ & $1.27+0.09$ & $1.00+0.09$ & $0.88+0.10$ & $0.93+0.12$  \\
\hline
0.025 & $88.65+0.00$ & $4.97+0.26$ & $1.39+0.08$ & $0.93+0.06$ & $\bf 0.86+0.04$ & $0.91+0.03$  \\
\hline
0.05  & $88.65+0.00$ & $88.65+0.00$ & $2.03+0.10$ & $1.02+0.05$ & $\bf 0.87+0.06$ & $\bf 0.82+0.02$  \\
\hline
0.1   & $88.65+0.00$ & $88.65+0.00$ & $3.23+0.27$ & $1.35+0.05$ & $\bf 0.87+0.03$ & $0.90+0.00$  \\
\hline
0.2   & $88.65+0.00$ & $88.65+0.00$ & $88.65+0.00$ & $2.12+0.11$ & $1.00+0.04$ & $0.93+0.08$  \\
\hline
0.4   & $88.65+0.00$ & $88.65+0.00$ & $88.65+0.00$ & $88.65+0.00$ & $1.86+0.05$ & $1.04+0.05$  \\
\hline
0.8   & $90.20+0.00$ & $88.65+0.00$ & $88.65+0.00$ & $88.65+0.00$ & $88.65+0.00$ & $1.60+0.09$ \\
\hline
$\alpha_k$ \textbackslash $\text{   } {|\mathcal{B}_k|}$ & 128 & 256 & 512 & 1024 & & \\
\hline
0.00625 & $1.14+0.02$ & $1.35+0.01$ & $1.56+0.03$ & $2.02+0.07$ & &  \\
\hline
0.0125 & $1.11+0.06$ & $1.17+0.09$ & $1.36+0.02$ & $1.68+0.03$ & &  \\
\hline
0.025 & $0.95+0.09$ & $1.07+0.08$ & $1.15+0.06$ & $1.40+0.05$ & &  \\
\hline
0.05  & $0.90+0.06$ & $1.01+0.09$ & $1.07+0.02$ & $1.22+0.05$ & &  \\
\hline
0.1   & $0.90+0.02$ & $0.92+0.07$ & $1.06+0.03$ & $1.13+0.02$ & &  \\
\hline
0.2   & $0.90+0.05$ & $0.96+0.08$ & $0.89+0.12$ & $1.11+0.09$ & &  \\
\hline
0.4   & $\bf 0.77+0.04$ & $0.92+0.07$ & $0.89+0.08$ & $0.92+0.06$ & &  \\
\hline
0.8   & $0.96+0.07$ & $0.90+0.10$ & $0.93+0.02$ & $0.98+0.05$ & & \\
\hline
\multicolumn{7}{|c|}{Dynamic Sampling Alternatives}\\
\hline
& s16-to-64 & s16-to-256  & s32-to-128 & s32-to-512 & s128-to-512 & s512-to-32 \\
\hline
0.1  & $1.07+0.06$ & $0.98+0.06$ & $\bf 0.79+0.10$ &  $0.93+0.02$ & $0.88+0.02$ & $0.95+0.05$ \\
\hline
& s16-to-64-MS & s16-to-256-MS & s32-to-128-MS & s32-to-512-MS & s128-to-512-MS & \\
\hline
0.1  & $0.90+0.12$ & $0.93+0.08$ & $\bf 0.73+0.08$ &  $1.23+0.10$ & $0.91+0.04$ &  \\
\hline
\end{tabular}}
\end{center}
\end{table*}

\end{document}